\documentclass[acmsmall]{acmart}

\AtBeginDocument{%
  \providecommand\BibTeX{{%
    \normalfont B\kern-0.5em{\scshape i\kern-0.25em b}\kern-0.8em\TeX}}}

\setcopyright{acmlicensed}
\copyrightyear{2024}
\acmYear{2024}
\acmDOI{10.1145/3696459}

\begin{document}

\title{Beware of Words: Evaluating the Lexical \textcolor{black}{Diversity} of Conversational \textcolor{black}{LLMs using ChatGPT as Case Study}}


\author{Gonzalo Mart\'inez}
\email{gonzmart@pa.uc3m.es}
\author{Jos\'e Alberto Hern\'andez}
\email{jahgutie@it.uc3m.es}
\affiliation{%
  \institution{Universidad Carlos III de Madrid}
  \streetaddress{Avda de la Universidad 30}
  \city{Legan\'es}
  \state{Madrid}
  \country{Spain}
  \postcode{28911}
}

\author{Javier Conde}
\email{javier.conde.diaz@upm.es}
\author{Pedro Reviriego}
\email{pedro.reviriego@upm.es}
\affiliation{
  \institution{Universidad Polit\'ecnica de Madrid}
  \streetaddress{Avda Complutense 30}
  \city{Madrid}
  \state{Madrid}
  \country{Spain}
  \postcode{28040}
}

\author{Elena Merino}
\email{elena.merino.gomez@uva.es}
\affiliation{
  \institution{Universidad de Valladolid}
  \streetaddress{Paseo del Cauce 59}
  \city{Valladolid}
  \state{Valladolid}
  \country{Spain}
  \postcode{47011}
}


\begin{abstract}
  The performance of conversational Large Language Models (LLMs) in general, and of ChatGPT in particular, is currently being evaluated on many different tasks, from logical reasoning or maths to answering questions on a myriad of topics. Instead, much less attention is being devoted to the study of the linguistic features of the texts generated by these LLMs. This is surprising since LLMs are models for language, and understanding how they use the language is important. Indeed, conversational LLMs are poised to have a significant impact on the evolution of languages as they may eventually dominate the creation of new text. This means that for example, if conversational LLMs do not use a word it may become less and less frequent and eventually stop being used altogether. Therefore, evaluating the linguistic features of the text they produce and how those depend on the model parameters is the first step toward understanding the potential impact of conversational LLMs on the evolution of languages. In this paper, we consider the evaluation of the lexical \textcolor{black}{diversity} of the text generated by LLMs \textcolor{black}{in English} and how it depends on the model parameters. A methodology is presented and used to conduct a comprehensive evaluation of lexical \textcolor{black}{diversity} using ChatGPT as a case study. The results show how lexical \textcolor{black}{diversity} depends on the version of ChatGPT and some of its parameters, such as the presence penalty, or the role assigned to the model. The dataset and tools used in our analysis are released under open licenses with the goal of drawing much-needed attention to the evaluation of the linguistic features of LLM-generated text. 
\end{abstract}



\received[accepted]{06 September 2024}

\maketitle

\section{Introduction}

The introduction of ChatGPT in 2022 triggered an exponential adoption of Large Language Model (LLM) based artificial intelligence tools. This has further accelerated the development of LLMs which is now a top priority for hyperscalers as shown by the recent introduction of Gemini \cite{geminiteam2023gemini} by Google or Grok\footnote{\url{https://x.ai/model-card/}} by xAI. These base LLMs are then fine-tuned to obtain conversational LLMs, such as ChatGPT, that can answer questions and follow instructions. As more and more LLM-based tools become available, understanding their performance becomes critical; for example, to select the right model for a given problem or to assess if a tool can be used for that problem in the first place. Measuring the performance on different tasks is also helpful in identifying limitations of the current models to focus efforts for improvements in the next versions of the models or in newer models developed from scratch. 

The evaluation of conversational LLMs and more broadly of Artificial Intelligence (AI) systems is challenging \cite{AIMeasurements}. A large number of benchmarks have been developed to evaluate the performance in a wide range of tasks and topics. For example, extensive test suites to evaluate how conversational LLMs solve mathematical problems are available \cite{Mathmeasuring} covering different subjects in maths with thousands of problems. Similarly, there are large benchmarks to evaluate conversational LLM knowledge in many different areas with multiple-choice questions. For example \cite{MultipleTasksmeasuring} includes 57 different tasks covering STEM, humanities, or social sciences. The number of tasks is extended to more than 200 in \cite{BIGMeasuring}. A similar approach is used in \cite{CommonSensemeasuring} to test common sense reasoning by choosing among different options to end a sentence. In these benchmarks, the results are quantified with a percentage of correct answers and models try to get as close as possible to 100\%\footnote{See for example Table 2 in \cite{geminiteam2023gemini}.}. 

Those elaborated benchmarks are capable of quantifying how conversational LLMs perform on different tasks with good accuracy. However, conversational LLMs are not only used to answer questions or to solve specific problems or games. They will be increasingly used to generate text; soon novels or textbooks written by conversational LLMs or with the help of LLMs will be common. Those texts will be read by humans and possibly used as well to train new LLMs thus influencing future writers both humans and LLMs \cite{martínez2023understanding}. In the case of generative AI models like LLMs, the use of AI-generated data for training can lead to performance issues \cite{degeneration, degenerationLLMs}. For humans, the texts they read condition language \textcolor{black}{learning} and use. Therefore, it is of interest to understand how LLMs use the language. For example, how do they use the vocabulary? If a word is not used by LLMs it will become less and less frequent and may eventually fall in disuse. 


These more subtle but fundamental questions are not adequately covered by existing conversational LLM benchmarks and performance evaluation procedures. Some works have considered the linguistic features of the text generated by LLMs \cite{HCC2, HCC3, reviriego2023playing} analyzing the presence of phonological biases and comparing some linguistic features or the lexical \textcolor{black}{diversity} of conversational LLMs and humans. However, to the best of our knowledge, there has been no attempt to understand how the linguistic features of conversational LLMs depend on the model parameters, the type of text generated, or the context and role given to the LLM. Similarly, there is no specific dataset that exercises conversational LLMs to produce different types of texts that can be used to evaluate the linguistic features of LLMs. In this paper we address both issues by 1) designing a simple dataset to exercise text generation in conversational LLMs and 2) using the dataset to evaluate how lexical diversity depends on the LLM parameters such as temperature or top probability and on the role assigned to the model \textcolor{black}{using ChatGPT as a case study}. The dataset and generated texts are publicly available so that they can be used in the first case to conduct the same evaluation for other conversational LLMs and in the second to enable other researchers to perform linguistic analysis of the conversational LLM-generated text.

The rest of the paper is organized as follows: section \ref{Evaluating} \textcolor{black}{discusses the concepts of lexical richness and lexical diversity and} identifies the requirements to evaluate the lexical richness of LLMs. Section \ref{Cave Verba} presents the proposed methodology for the evaluation that includes both the selection of prompts and parameters as well as the testing procedure. \textcolor{black}{The results obtained when applying the proposed methodology to the ChatGPT case study are presented in section \ref{Results} discussing the main findings.} The paper ends with the conclusions and a discussion of future work.

\section{Evaluating the Lexical Richness of Large Language Models}
\label{Evaluating}

\textcolor{black}{Lexical richness aims to evaluate an individual's command of vocabulary, encompassing both the breadth, represented by the number of words or vocabulary size, and the depth, indicating how thoroughly those words are understood. This includes lexical density given by the number of content words in the text but also the lexical diversity which itself has several dimensions: variability, volume, evenness, rarity, dispersion, and disparity \cite{jarvis2013defining}. The level of sophistication of the vocabulary used is also of interest. In fact, the features, metrics, and terminology to study the lexical aspects of text is an active area of research.  To evaluate those features, different metrics have been proposed over the years that could be applied to LLM-generated text \cite{LexicalRichnessdiversityDefinition}.}

\textcolor{black}{In humans, the lexical features depend on different factors that introduce linguistic variations. Those include a context (diaphasic variation), social (diastratic variation) geographic (diatopic variation), and historic (diachronic variation) factors among others \cite{kastovsky2011history}. In the case of LLMs the lexical features of the text also depend on multiple factors.} For example, on the task that the LLM is performing; the text to answer a computing question can be very different from the one of a literary essay. A similar reasoning applies to the role or context that is given to the LLM, the text for the same task will be different if we ask the LLM to write it as a child or as a sophisticated writer. Those two factors are similar to what happens with humans, the text we produce depends on the task and person. LLMs also have other configuration parameters that are directly related to the model itself, for example, temperature or top probability. Therefore, evaluating the lexical features or more broadly the linguistic features of LLMs is a complex task that requires careful analysis to design a methodology that can capture all those factors.   
\textcolor{black}{To design such a methodology, it is also important to understand what are the potential goals when it is used for evaluation, which include: }

\begin{itemize}
    \item  Understanding the impact of the different factors (task, role, parameters) on the lexical features of the generated text for a given model.
    \item  Comparing the lexical features of the text generated by different LLMs. 
\end{itemize}

\textcolor{black}{The first potential goal focuses on a single LLM and will enable, for example, understanding and quantifying the lexical diversity versus the temperature of the model.} As of today, we know that high-temperature values make the text more creative and diverse, but how much? Is the lexical \textcolor{black}{diversity} task or role-dependent? The second \textcolor{black}{potential} objective is of interest to compare the lexical features as LLMs evolve, for example, how do the lexical features change from ChatGPT3.5 to ChatGPT4? 


An evaluation of the lexical features of LLMs should ideally include:

\begin{enumerate}
    \item \textbf{A large corpus of prompts} for each task to ensure that the results are representative of the text generated by the model.
    \item \textbf{Different tasks and topics} to test the model when writing different types of text on different topics.
    \item \textbf{Different roles} to evaluate the importance of the role and context on the generated text.
    \item \textbf{Different hyperparameter configurations} to assess the impact of each parameter on the generated text.
\end{enumerate}

In the next section, we propose an initial test suite to evaluate the lexical features of LLMs that tries to capture all those features focusing on the evaluation of ChatGPT.  

\section{Cave Verba: Beware of Words}
\label{Cave Verba}

In this section, we describe the methodology used to evaluate the lexical diversity of ChatGPT that we denote as ``Cave Verba'' a Latin expression coined on the famous Pompeian ``cave canem''\footnote{Beware of the dog.}. ``Cave Verba'' can be translated as ``beware of words'', which is intended to highlight the importance of words in AI-generated texts but also to stress the carefulness needed when studying its outputs. First, we describe the test suite, and then the testing procedure. The test prompts, as well as the data generated, are shared\footnote{The data for the TOEFL prompts is not shared as it is used for commercial purposes.} in a public repository\footnote{\url{https://doi.org/10.5281/zenodo.11121394}} so that further linguistic analysis can be made on the text generated by ChatGPT with different configurations and parameter values.

\subsection{Test suite}

The tests are intended to provide a comprehensive evaluation of the lexical diversity at a reasonable computational cost. In the following, tasks, roles, and parameters used in the test are described.

\subsubsection{Tasks and topics}

The first part of the test suite is to define the tasks and associated prompts. Since we are interested in the features of LLM-generated text we focus on prompts that force the LLM to write new text and exclude summarization or translation tasks. In particular, we consider the following tasks:

\begin{enumerate}
    \item Essay writing: the LLM is asked to write a short essay about a given topic.
    \item Question answering: The LLM is asked to answer questions on different topics\footnote{Note that the LLM is not given a choice of answers to select one as in many existing benchmarks but rather asked to write a textual answer.}.     
\end{enumerate}

For essay writing two different sets of prompts are used. The first one corresponds to TOEFL essays\footnote{\url{https://github.com/rexshijaku/chatgpt-generated-text-detection-corpus/blob/main/full_texts/questions.txt}} and the second to topics compiled by the New York Times\footnote{\url{https://www.nytimes.com/2023/07/19/learning/175-writing-prompts-to-spark-discussion-and-reflection.html}} for argumentative and narrative writing. For question answering, we use data from the HC3 dataset\footnote{\url{https://huggingface.co/datasets/Hello-SimpleAI/HC3/}}. In particular, subsets of 40 questions on medicine, finance, computing, open, and Reddit are randomly selected from those categories in the HC3 dataset\footnote{Several such subsets were tested for each category with a few model settings to check that they produce similar lexical \textcolor{black}{diversity} metrics.}. 

A summary of the data used is given in Table \ref{tab:tasks}. It can be seen that we have a large number of prompts that are distributed on different tasks and topics and come from different sources to ensure that results are representative. In total, over 500 prompts which is a reasonable size when taking into account that prompts have to be tested for many different parameter values.

\begin{table}
  \caption{Tasks testing data}
  \label{tab:tasks}
  \begin{tabular}{cccc}
    \toprule
    Task &  Category & Number of prompts\\
    \midrule
    Essay writing & TOEFL & 126\\
    Essay writing & NYT argumentative  & 57\\
    Essay writing & NYT narrative  & 118\\    
    \midrule
    Question answering & open & 40 \\
    Question answering & finance & 40\\
    Question answering & medicine & 40\\
    Question answering & Reddit & 40 \\
    Question answering & computing & 40\\
  \bottomrule
\end{tabular}
\end{table}

\subsubsection{Roles}

To assess the impact of the role assigned to the LLM on lexical \textcolor{black}{diversity}, we have selected a set of roles that are expected to be different in terms of their use of the language and, in particular, of its lexicon. The effect of social parameters on language is studied by sociolinguistics \cite{Sociolinguistics}. For example, social class, gender, age, educational level, or ethnicity can influence the use of the language at different levels: phonetic, morphological, syntactic, lexical-semantic, and textual \cite{labov1972sociolinguistic}. The number of sociolinguistic investigations conducted in the field of written production is much lower compared to those related to oral production \cite{Write_Oral}, and often, English is not used as the experimental language. Nevertheless, we believe that their general findings can be extrapolated for the purposes of an initial comparative analysis with the results of AI production through role assignment by considering age, social class, and gender.  

Regarding social class, it is expected that speakers from higher classes produce texts with a richer lexicon than those from lower classes \cite{Lexical_Richness_Class}. In terms of age, adults have a more extensive vocabulary than children \cite{Lexical_Richness_Age}, and the effects of aging may influence lexical impoverishment \cite{Lexical_Richness_Age1, Lexical_Richness_Age2}. Differences in linguistic production based on gender are the subject of recent research and exhibit a highly heterogeneous nature, in \cite{Lexical_Richness_Gender} after analyzing the oral production of a group of 17 women and 13 men, it was found that male discourse was richer with generally longer sentences. In contrast, female discourse contained shorter sentences with more lexical repetitions. In broader spectrum studies, gender differences were found to be irrelevant \cite{Lexical_Richness_Age}.  Other studies have found only slight gender differences in linguistics features in formal and informal contexts, such as word length, which is greater in male subjects, or the number of adjectives used, with a higher usage observed in female groups \cite{Lexical_Richness_Gender1}.

Based on these studies, we have selected eleven roles for evaluation:

\begin{enumerate}
    \item Default: No role is assigned to the LLM, the default role is used.
    \item Child: The LLM is requested to answer as a five-year-old child.    
    \item Young adult male: The LLM is requested to answer as a young male adult.    
    \item Young adult female: The LLM is requested to answer as a young female adult.    
    \item Elderly adult male: The LLM is requested to answer as an elderly male adult.    
    \item Elderly adult female: The LLM is requested to answer as an elderly female adult.    
    \item Affluent adult male: The LLM is requested to answer as an affluent male adult.    
    \item Affluent adult female: The LLM is requested to answer as an affluent female adult.    
    \item Lower-class adult male: The LLM is requested to answer as a lower-class male adult.    
    \item Lower-class adult female: The LLM is requested to answer as a lower-class female adult.    
    \item Erudite\footnote{The term erudite is used to try to make the LLM answer as a highly educated native user of the language.}: The LLM is requested to answer as an erudite who uses a rich vocabulary.

\end{enumerate}

These roles are intended to provide information on how age, gender, and social class factors influence the lexical \textcolor{black}{diversity} of AI-generated text.

\subsubsection{LLM parameters}

The hyperparameters of each LLM are different and thus the testing has to be adjusted for each model. However, the differences are in many cases small and some parameters are common to most LLMs. Therefore a similar approach can be used for the testing. In our evaluation, we focus on OpenAI conversational tools: the well-known ChatGPT which uses as base different LLMs, such as GPT4 \cite{openai2023gpt4}. ChatGPT is a commercial product and not many details are provided on how the base LLM is modified or complemented, additionally, the API provides limited visibility into the model results and thus our evaluation approach is to consider ChatGPT as a black box. The main model parameters that are accessible through their API\footnote{\url{https://platform.openai.com/docs/api-reference}} are described by OpenAI as follows:

\begin{enumerate}
    \item Temperature: ``Higher values like 0.8 will make the output more random, while lower values like 0.2 will make it more focused and deterministic.''
    \item Top probability: ``An alternative to sampling with temperature, called nucleus sampling, where the model considers the results of the tokens with top\_p probability mass. So 0.1 means only the tokens comprising the top 10\% probability mass are considered.''
    \item Frequency penalty: ``Number between -2.0 and 2.0. Positive values penalize new tokens based on their existing frequency in the text so far, decreasing the model's likelihood to repeat the same line verbatim.''  
    \item Presence penalty: ``Number between -2.0 and 2.0. Positive values penalize new tokens based on whether they appear in the text so far, increasing the model's likelihood to talk about new topics.''
\end{enumerate}

From their descriptions, it seems that all four parameters can affect the lexical \textcolor{black}{diversity} of the generated text and should be evaluated. The parameters are also quite general and similar parameters are available in other LLMs like Llama \cite{LLAMA2} or Mistral \cite{Mistral}. Temperature is a classical parameter to control how probabilities are assigned in the Softmax normalization that maps values from the last layer of the model to the probabilities of each token being selected \cite{Temperature}. The top probability or top\_p is also a common parameter in LLMs and restricts the choice to the tokens that concentrate most of the probability, so the sampling is done from the most likely tokens only \cite{NucleusSampling}. The last two parameters try to avoid repetitions in the text by introducing a penalty for frequently used tokens or for all tokens that have been already used in the generated text, similar parameters are common in other LLMs too. After describing each of the parameters, it seems that all four are likely to have an impact on the lexical \textcolor{black}{diversity} of the generated text.

\subsection{Test procedure}

The first step in the evaluation is to generate the text with the AI model. This is done with a script that makes calls to OpenAI's models with the desired parameter values and context and iterates over all the prompts in the different tasks and topics to produce the result files with the text. Those files are then processed, and the relevant lexical \textcolor{black}{diversity} metrics are computed.

The LexicalRichness module \cite{lexRichMetrics} is then used to compute the metrics for each of the categories in each task. Four metrics are selected for the comparisons. Two of them are based on the types and tokens over the entire set of texts for the category, and two are computed for each of the texts individually and then averaged to report the mean value. The first two metrics are intended to capture the lexical \textcolor{black}{diversity} over the entire set of answers for each category. The other two metrics focus on measuring the lexical \textcolor{black}{diversity} of each of the answers and then take the average per category.  

In more detail, the Root Type-Token Ratio (RTTR) and the Maas metric are calculated on the texts of a given category \cite{MetricsLexical1, MetricsLexical2}. These metrics are based on the total number of words ($nTokens$) and the number of distinct words ($nTypes$) and introduce modifications over the plain Type-Token Ratio (TTR) metric to make them independent of the number of words in the text. In particular:

\begin{equation}
    RTTR = \frac{nTypes}{\sqrt{nTokens}}
\end{equation}

\begin{equation}
    Maas = \frac{\log(nTokens)-\log(nTypes)}{[\log(nTokens)]^2}
\end{equation}

For example, for the TOEFL category and default settings, the 126 essays generated by ChatGPTv3.5 have in total 52,778 words and 4,551 distinct words which are used as the token and types on the calculations of the RTTR and Maas metrics, with values of 19.8 and 0.02 respectively. Therefore, with these two metrics the diversity of the lexicon used across all the TOEFL essays is evaluated.

Additionally, two alternative metrics are computed for each of the texts: the Moving average TTR  (MATTR) \cite{MATTR} and the Measure of Lexical Diversity (MTLD) \cite{MTLD}. These metrics operate sequentially on the text considering blocks or segments. For example, MTLD counts words until the TTR reaches a predefined threshold and then starts a new segment, and so on. For these metrics, since the answers are independent merging them and computing the metric on the entire set can create artifacts. Therefore, it is more appropriate to evaluate the metric for each of the texts individually and then report the average across all the texts in a given category. Therefore these metrics evaluate the lexical \textcolor{black}{diversity} of each essay individually and independently from the rest of the essays in the same category indicating the lexical \textcolor{black}{diversity} of each essay in isolation.

Finally, the trends and relative results are similar when the texts are preprocessed to keep only the lemmas or when removing the most common stop words. Therefore, the observed dependencies of lexical \textcolor{black}{diversity} on the model's parameters and context seem to be independent of the lemmatization or preprocessing done on the texts.


\textcolor{black}{\section{Case study}}
\label{Results}

\textcolor{black}{To illustrate the use of the proposed methodology, the evaluation of the lexical diversity of the text generated by ChatGPT is used as a case study. The next subsection presents the results, then the analysis and insights are discussed in the second subsection.}

\subsection{\textcolor{black}{Evaluation results}}

As discussed before, our goal in these experiments is to understand the impact of the model parameters on lexical \textcolor{black}{diversity}. To that end, we have run the entire dataset of prompts for ChatGPT3.5\footnote{The exact models used in the evaluation are 'gpt-3.5-turbo-1106' and 'gpt-4-1106-preview'}. For ChatGPT4, we have selected the NYT dataset and run both argumentative and narrative prompts.

The results are shown in the following figures which are divided into two groups, one for the metrics that are computed per prompt (MATTR and MTLD) and another for the metrics that are computed for all the prompts in each category. In the first case, the metrics are computed per prompt and the average for all prompts in each category is reported as discussed in the previous section. \textcolor{black}{The 95\% confidence interval for the average was calculated obtaining less than 1\% for MATTR and less than 5\% for MTLD in all configurations that produced correct texts.} The values used for each category are the following:

\begin{enumerate}
    \item Role: Default (D), child (C), young adult male (YAM), young adult female (YAF), elderly adult male (EAM), elderly adult male (EAF), affluent adult male (AAM), affluent adult female (AAF), lower class adult male (LAM), lower class adult female (LAF) and erudite (ERU).
    \item Temperature: {0, 0.2, 0.4, 0.6, 0.8, 1,0, 1.2, 1.4, 1.6, 1.8, 2.0}.
    \item Top\_p: {0, 0.1, 0.2, 0.3, 0.4, 0.5, 0.6, 0.7, 0.8, 0.9, 1}.
    \item Presence penalty: {-2.0, -1.5, -1, -0.5, 0, 0.5, 1, 1,5, 2.0}.
    \item Frequency penalty: {-2.0, -1.5, -1, -0.5, 0, 0.5, 1, 1,5, 2.0}.
\end{enumerate}

\begin{figure}[h]
  \centering
  \includegraphics[scale=0.45]{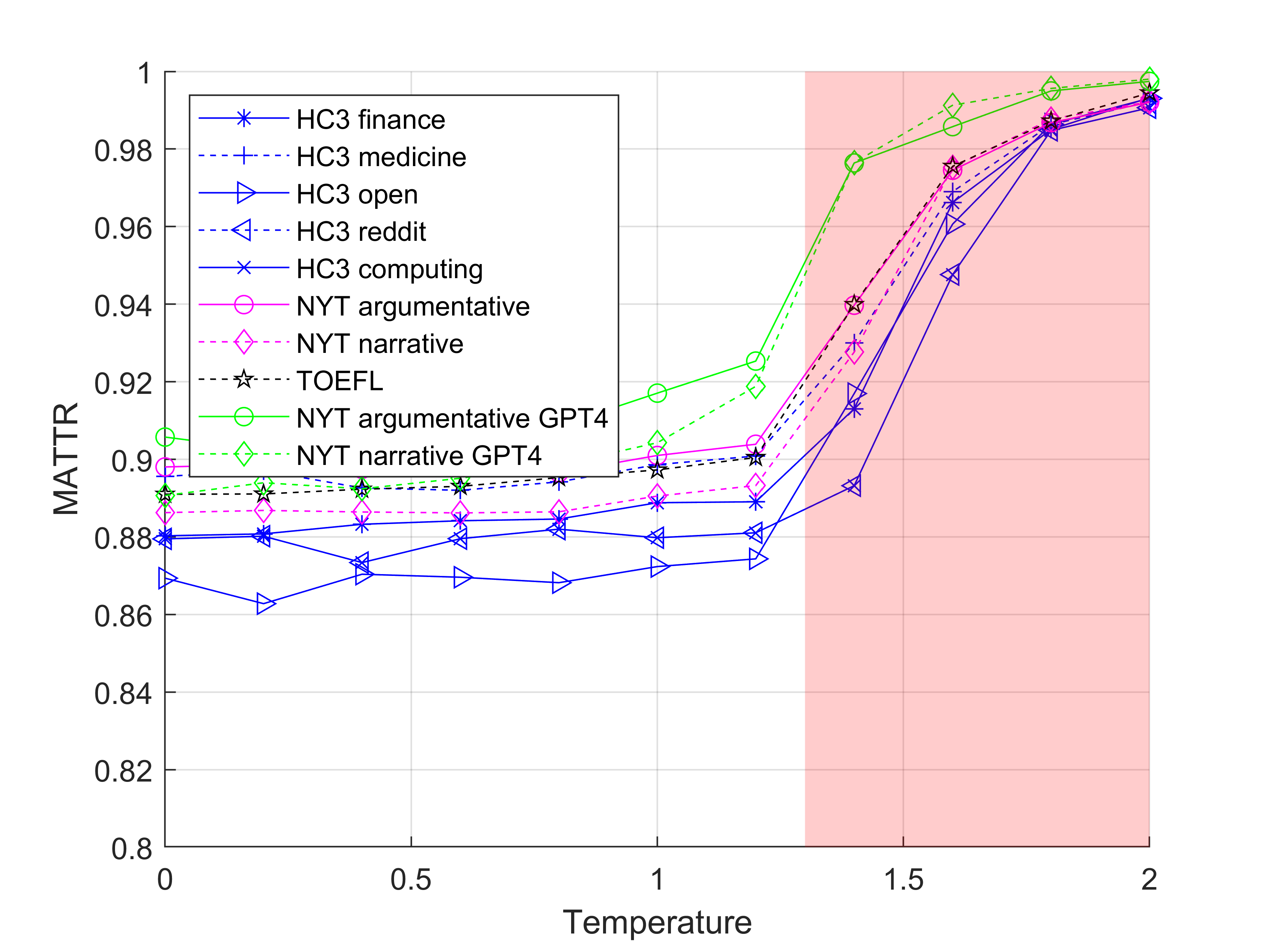}
  \includegraphics[scale=0.45]{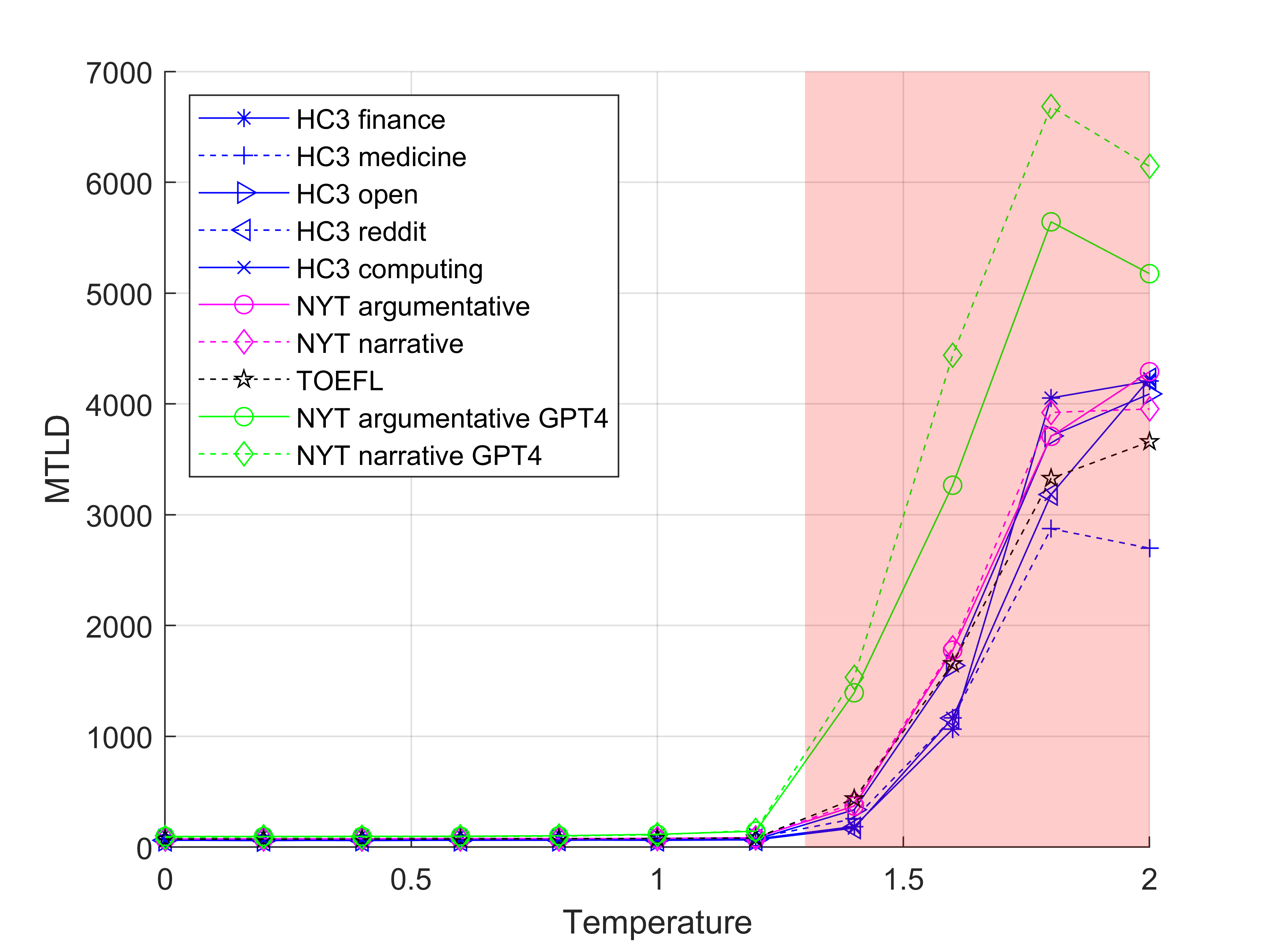}
  \caption{Lexical \textcolor{black}{diversity} per prompt metrics (MATTR, MTLD) average across all prompts versus Temperature, range values that produce incorrect results are marked in red.}
  \label{fig;T1}
\end{figure}

\begin{figure}[h]
  \centering
  \includegraphics[scale=0.45]{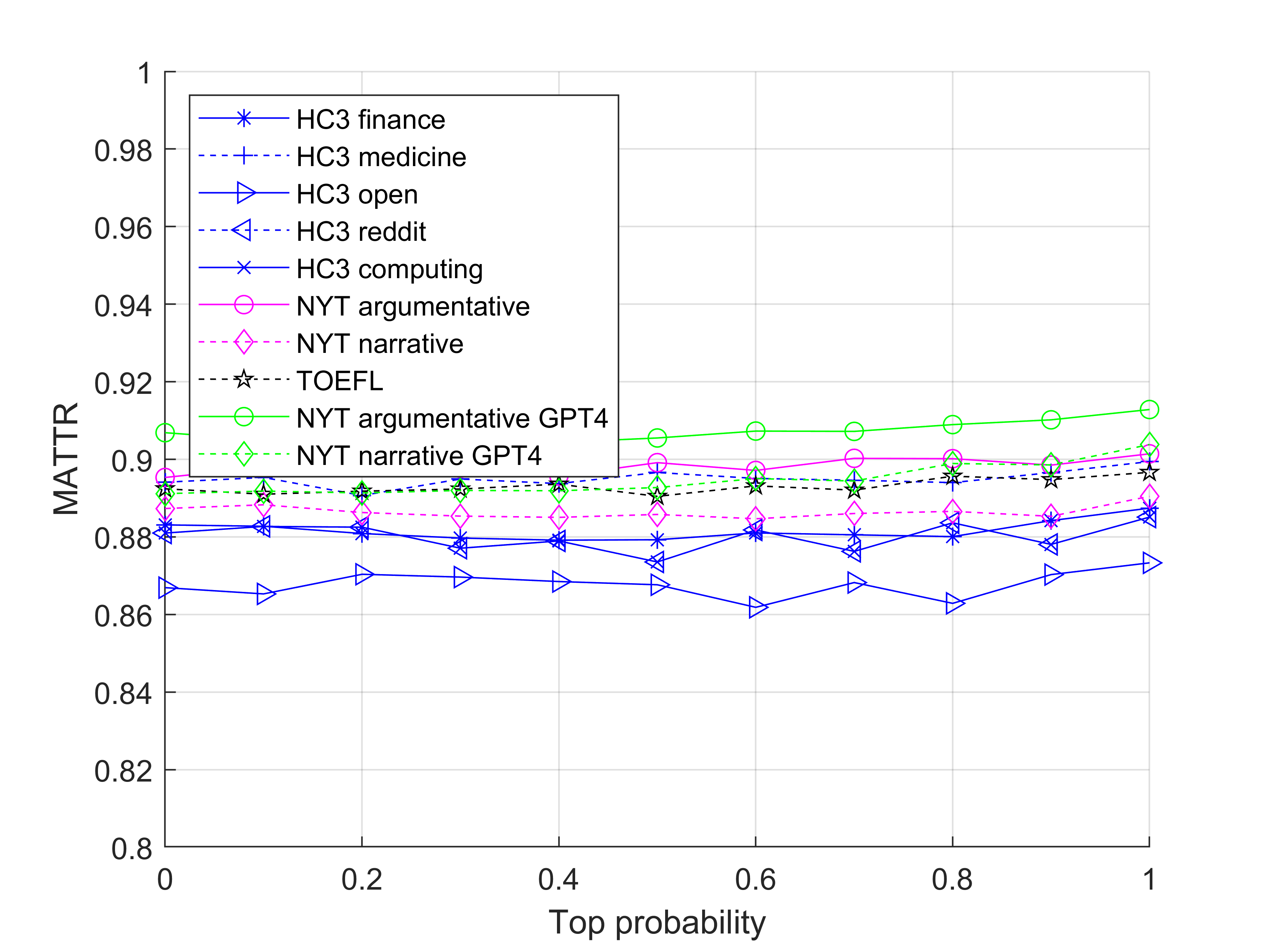}
  \includegraphics[scale=0.45]{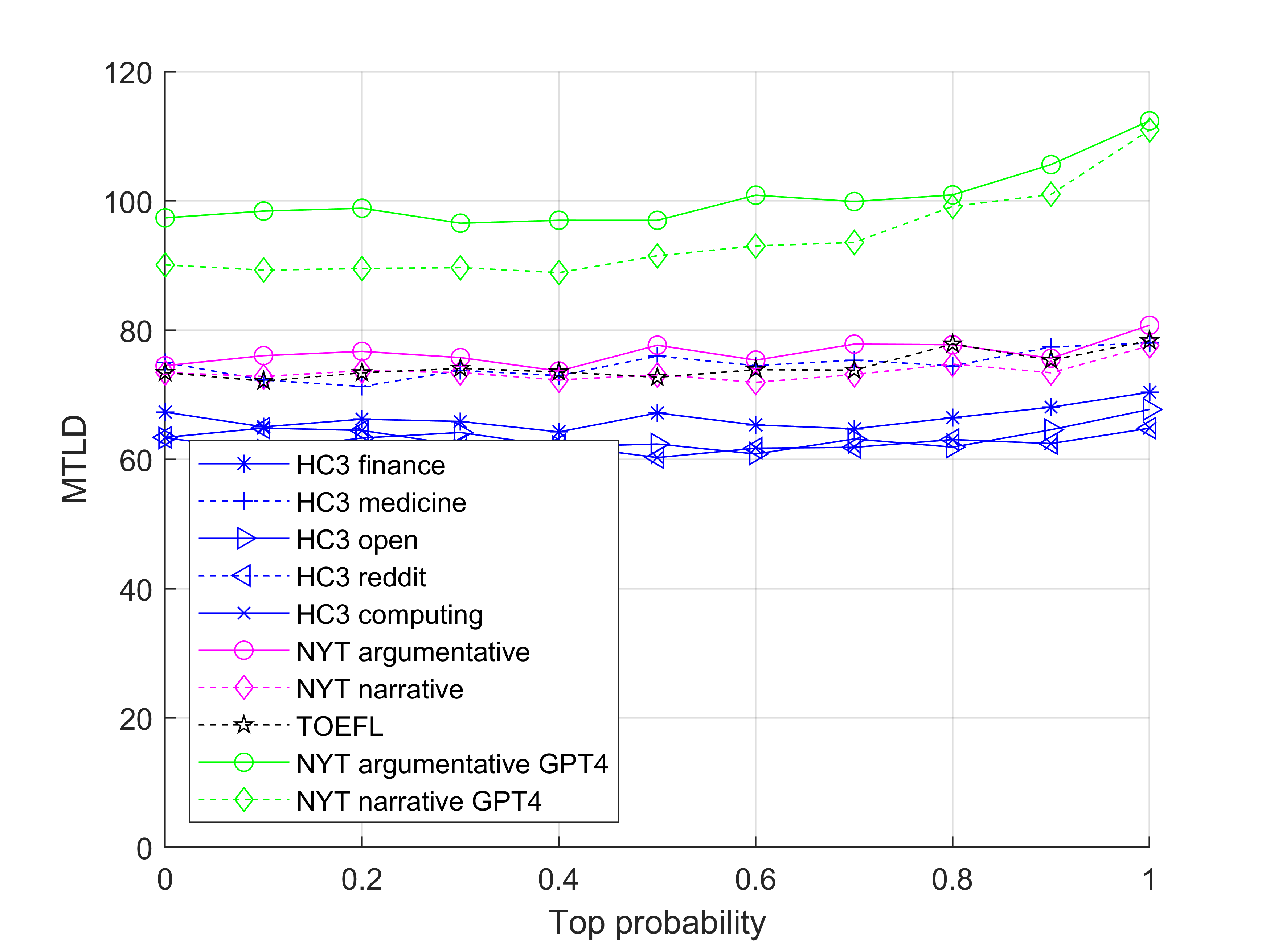}
  \caption{Lexical \textcolor{black}{diversity} per prompt metrics (MATTR, MTLD) average across all prompts versus Top Probability.}
  \label{fig:TP1}
\end{figure}

\begin{figure}[h]
  \centering
  \includegraphics[scale=0.45]{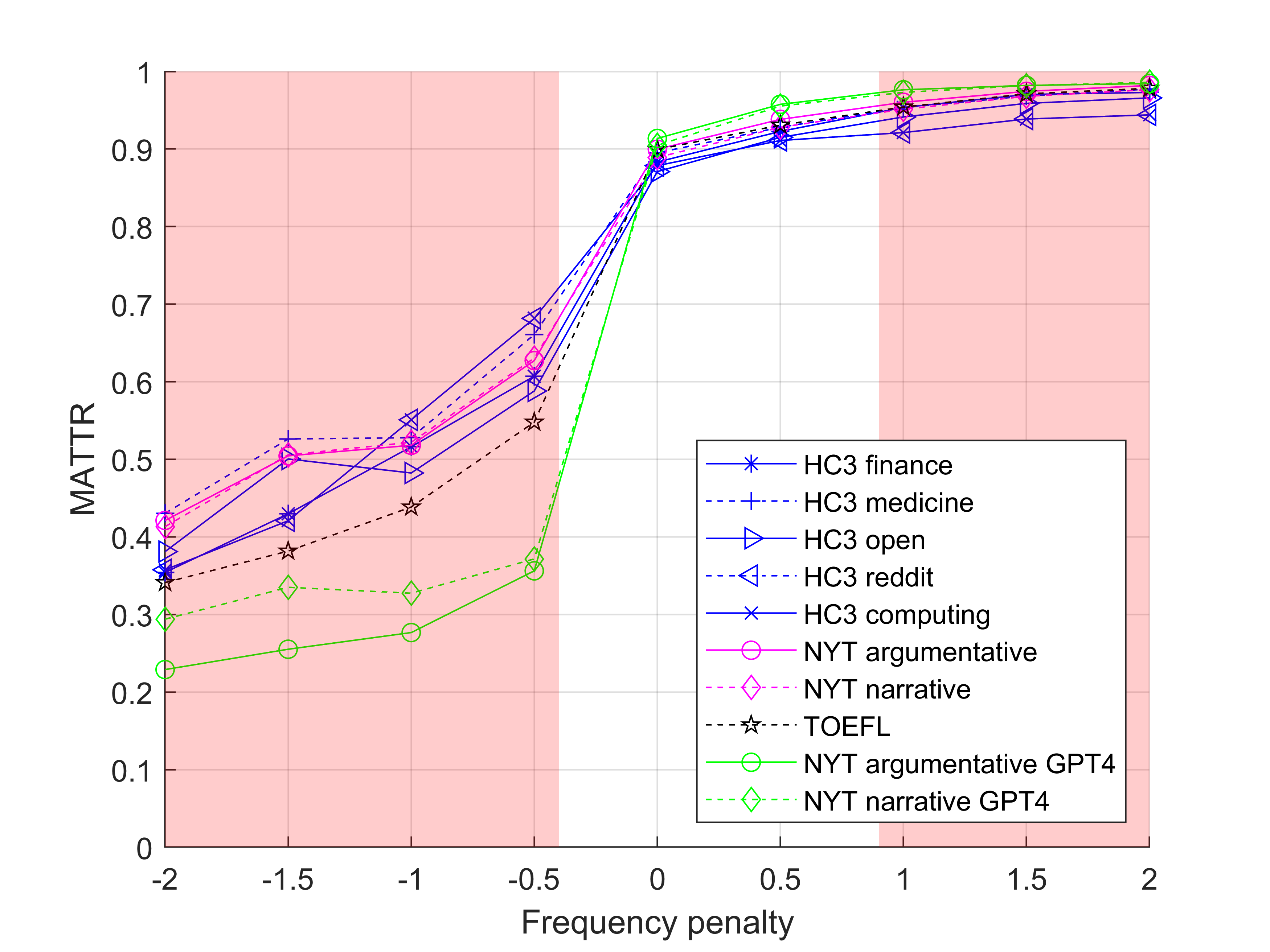}
  \includegraphics[scale=0.45]{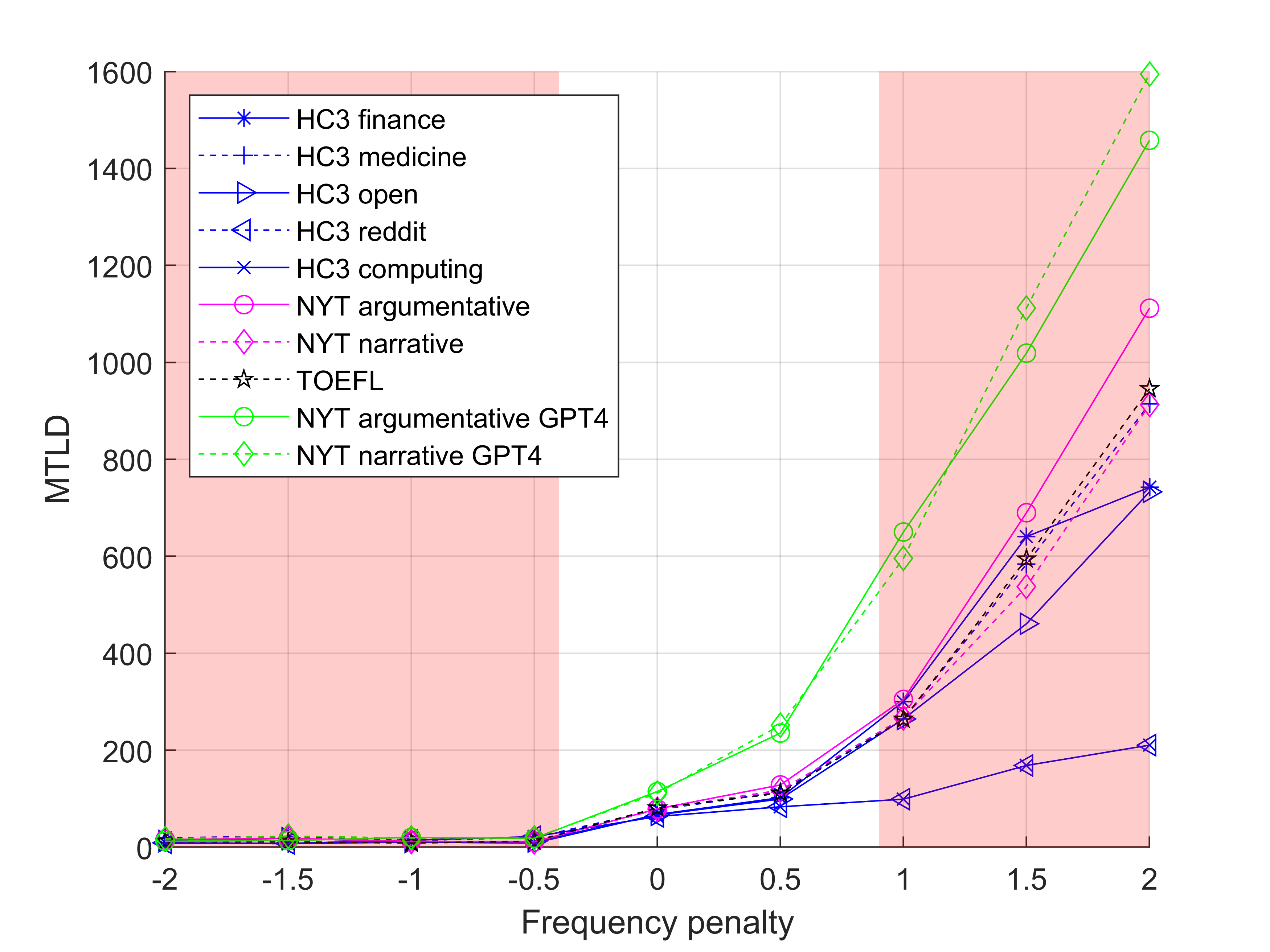}
  \caption{Lexical \textcolor{black}{diversity} per prompt metrics  (MATTR, MTLD) average across all prompts versus Frequency Penalty, range values that produce incorrect results are marked in red.}
  \label{fig:F1}
\end{figure}

\begin{figure}[h]
  \centering 
  \includegraphics[scale=0.45]{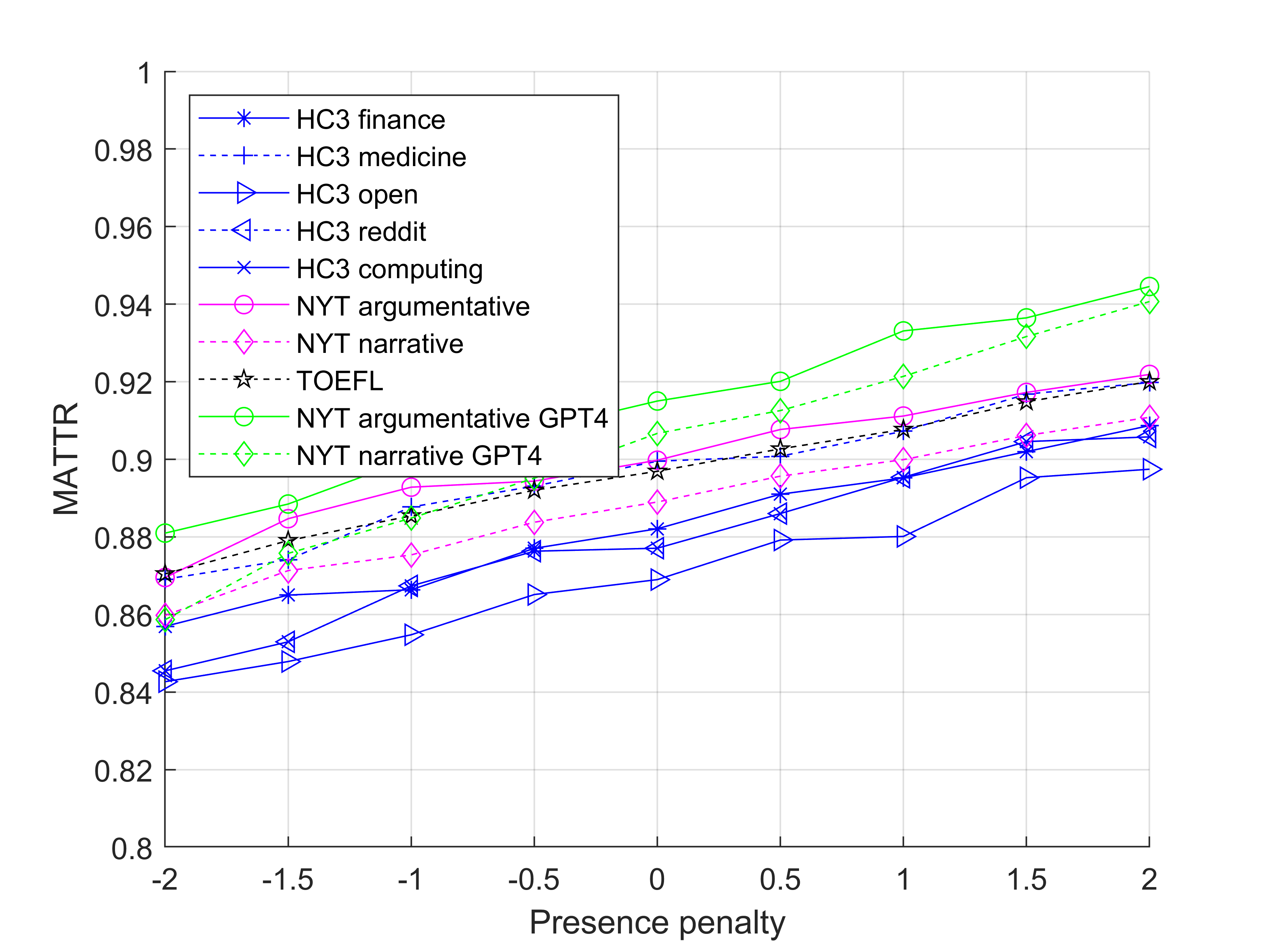}
  \includegraphics[scale=0.45]{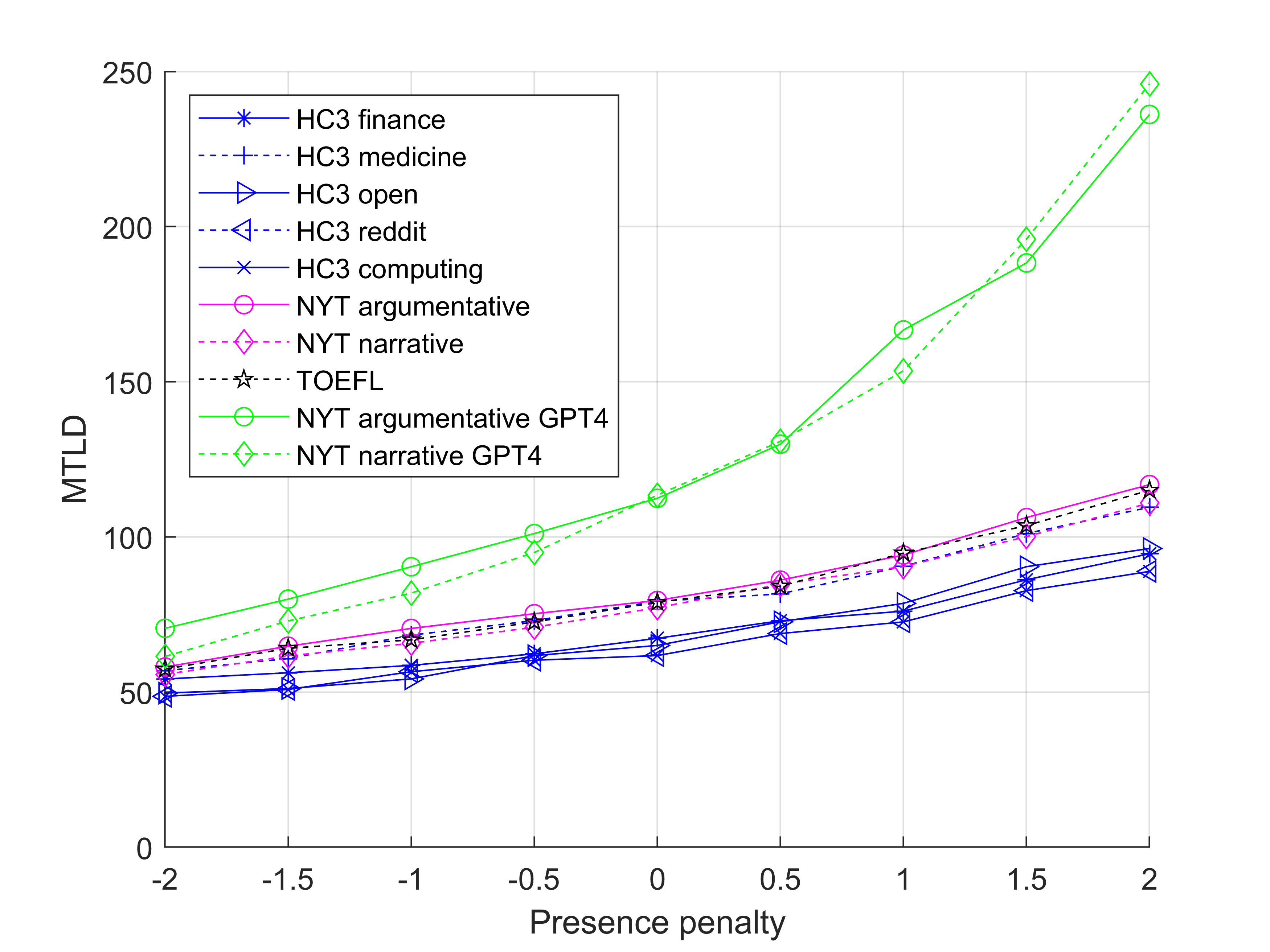}
  \caption{Lexical \textcolor{black}{diversity} per prompt metrics (MATTR, MTLD) average across all prompts versus Presence Penalty.}
  \label{fig:P1}
\end{figure}

The results for the first pair of metrics and the four model parameters are summarized in figures \ref{fig;T1},\ref{fig:TP1},\ref{fig:F1},\ref{fig:P1}. It can be seen that both MATTR and MTLD increase significantly with temperature and frequency penalty. Looking at the text generated for high values of both parameters (and for negative values of the frequency penalty), it is often incorrect with non-valid words and repetitions of letters and words. This occurs for both ChatGPT3.5 and ChatGPT4. In more detail when the temperature is equal to or larger than 1.4 or when the frequency penalty is equal to or smaller than -0.5 or equal to or larger than 1.0, the percentage of non-valid words is large so the text generated is not useful. The range values for which incorrect texts are generated are marked in red in the figures and not considered in the analysis. For top probability and presence penalty, this effect was not observed and texts were valid for all the values of these two parameters. 

The effects of high temperature or frequency penalty values are illustrated with two examples of excerpts from the model answers:
``Humans are social creatures, and we often seek thoughtful governance as appealing undertaking lengthy fulfilievedg bent structure ballo Furthermore recovering distresso well burAdditionallyilationrequested whenhigh.utilisation ret burden serieEdit teal'' or 
``In conclusion,itisdiretenely crucial renember preservesavingaraelemtery environmental nhaturalresourcesinfrom morleywalmon polutionseaLeverls.Designed implemented windows utility ultimate goal yayhd ashes reasons psothosglobaleyare''. It is clear that those parameter values should not be used. 

Considering only the range of values that produce valid texts, the temperature does not seem to affect MATTR and MTLD significantly, only a small increase is observed when raising the temperature. Similarly, for the frequency penalty, only two values (0 and 0.5) provide consistent and valid texts and the lexical \textcolor{black}{diversity} is slightly larger for the largest value (0.5). The top probability does not seem to have any relevant impact on the lexical \textcolor{black}{diversity} of the generated text except for ChatGPT4 which shows a small increment when the top probability is 1.0. Finally, the lexical \textcolor{black}{diversity} shows an approximately linear behavior with the presence penalty. The trends are consistent across all the tasks and categories. 

Looking at the lexical \textcolor{black}{diversity} of ChatGPT3.5 and ChatGPT4, for the same settings, ChatGPT4 has a larger \textcolor{black}{diversity} in most cases. The diversity is also generally larger for essay writing (TOEFL and NYT) than for question answering (HC3). 

\begin{figure}[h]
  \centering 
  \includegraphics[scale=0.45]{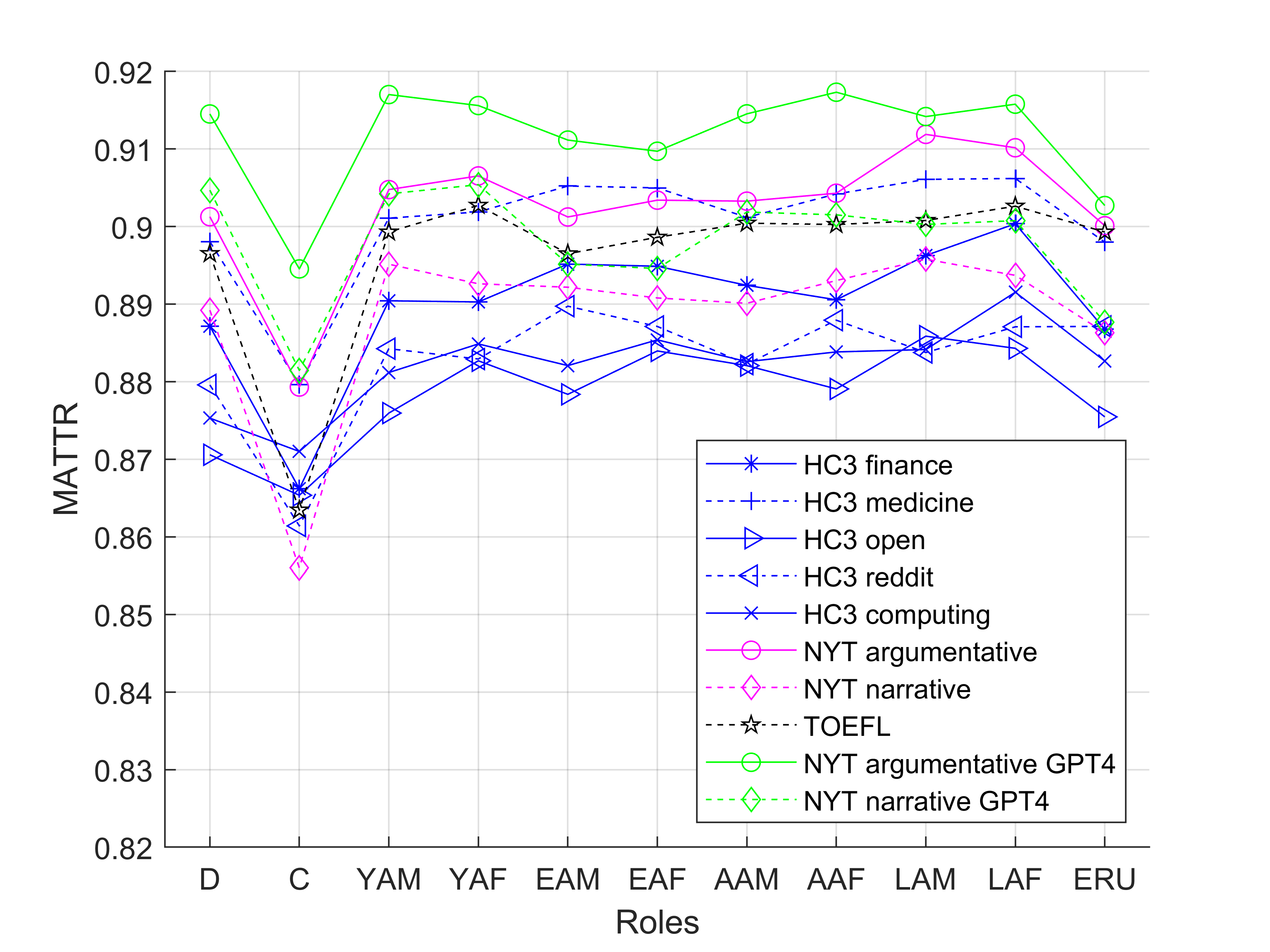}
  \includegraphics[scale=0.45]{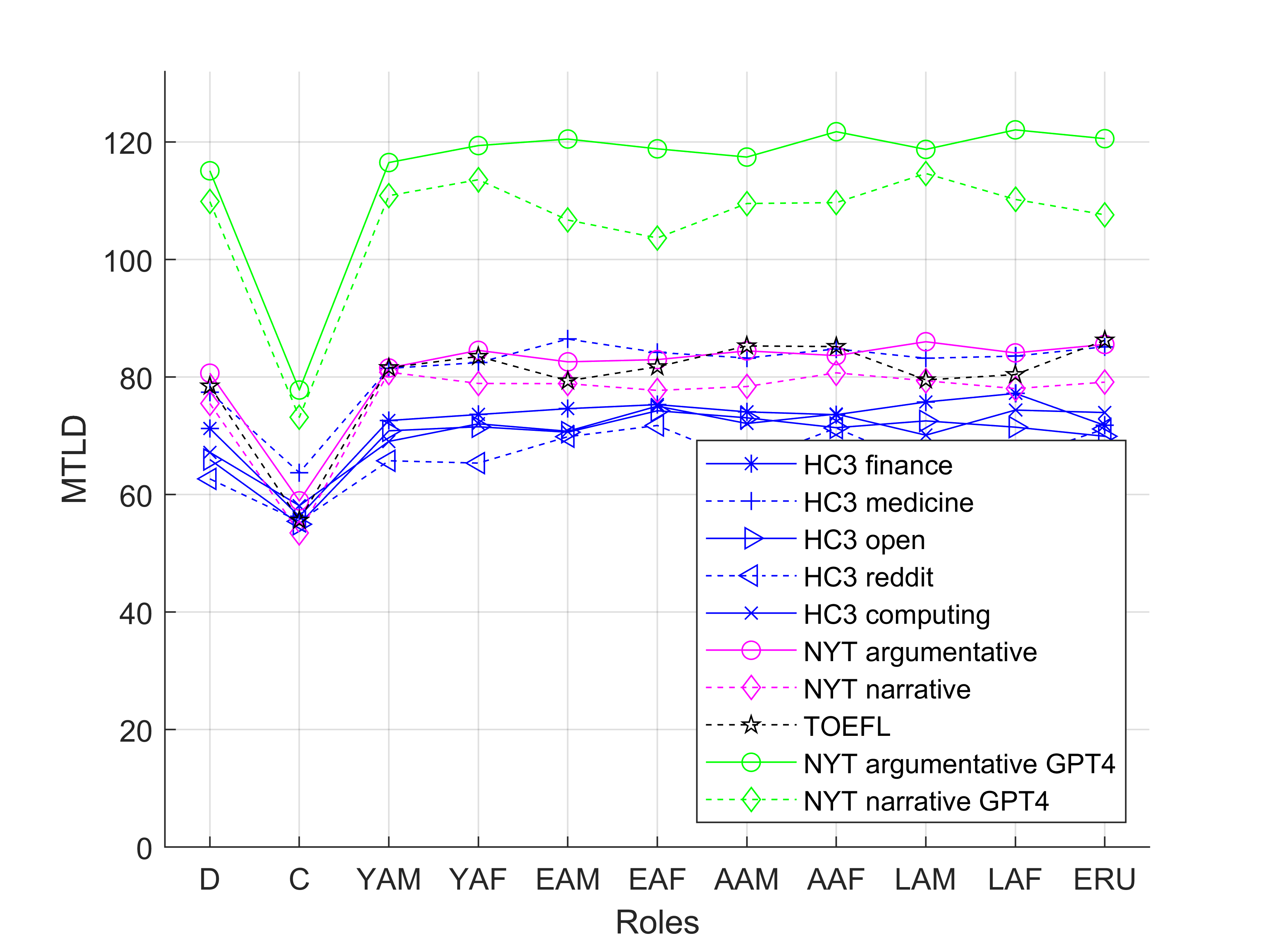}
  \caption{Lexical \textcolor{black}{diversity} per prompt metrics (MATTR, MTLD) average across all prompts for different roles.}
  \label{fig:R1}
\end{figure}

The results for the different roles on the first pair of metrics are shown in Figure \ref{fig:R1}. It can be observed that lexical \textcolor{black}{diversity} is lower for the ``child'' role while for the rest differences are small. It seems that age (excluding children), sex, or social class do not have a significant impact on the MATTR or the MTLD metrics. Comparing ChatGPT4 versus ChatGPT3.5, diversity tends to be larger for essay writing than for question answering as in previous results.

\begin{figure}[h]
  \centering
  \includegraphics[scale=0.45]{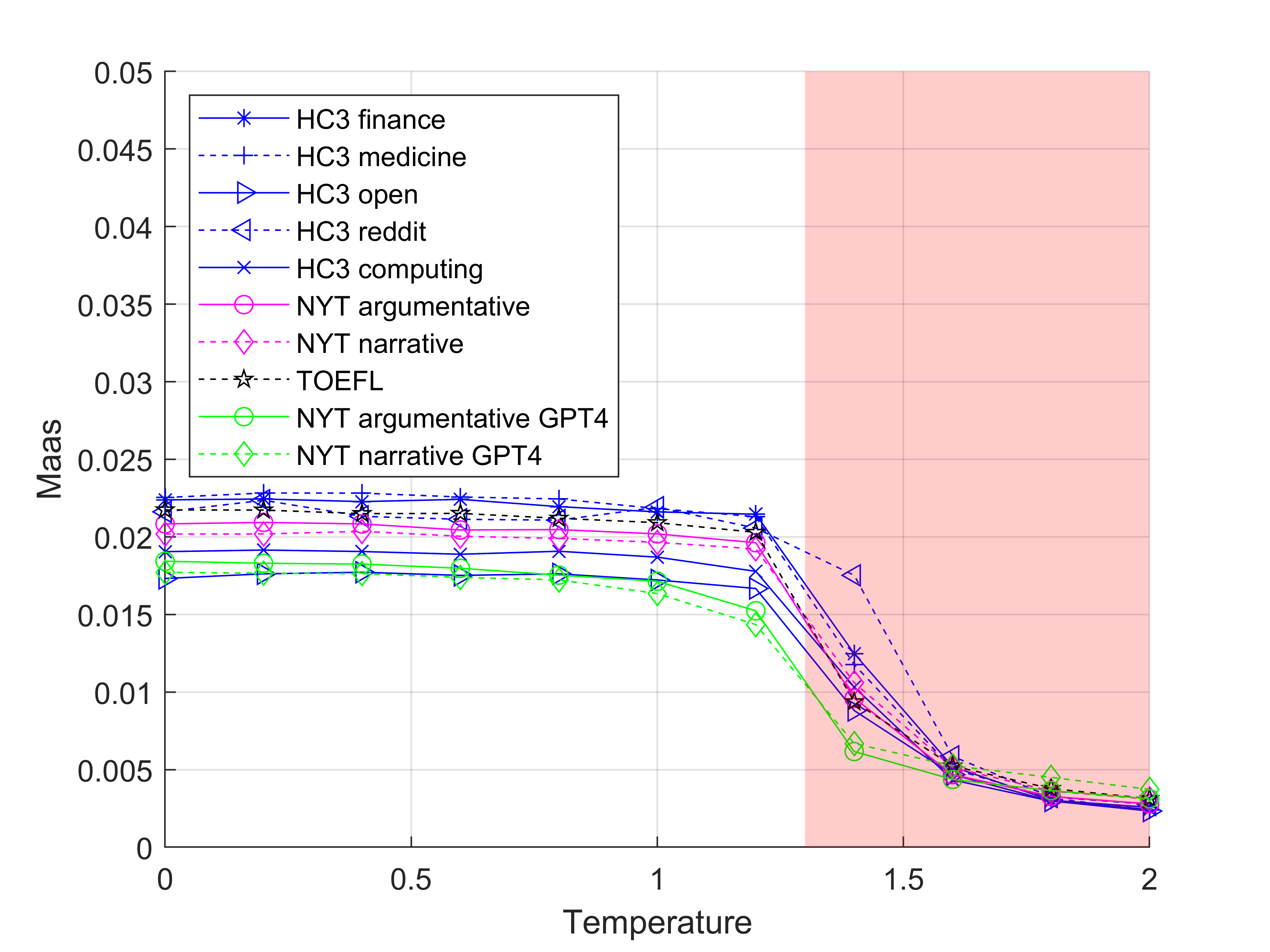}
  \includegraphics[scale=0.45]{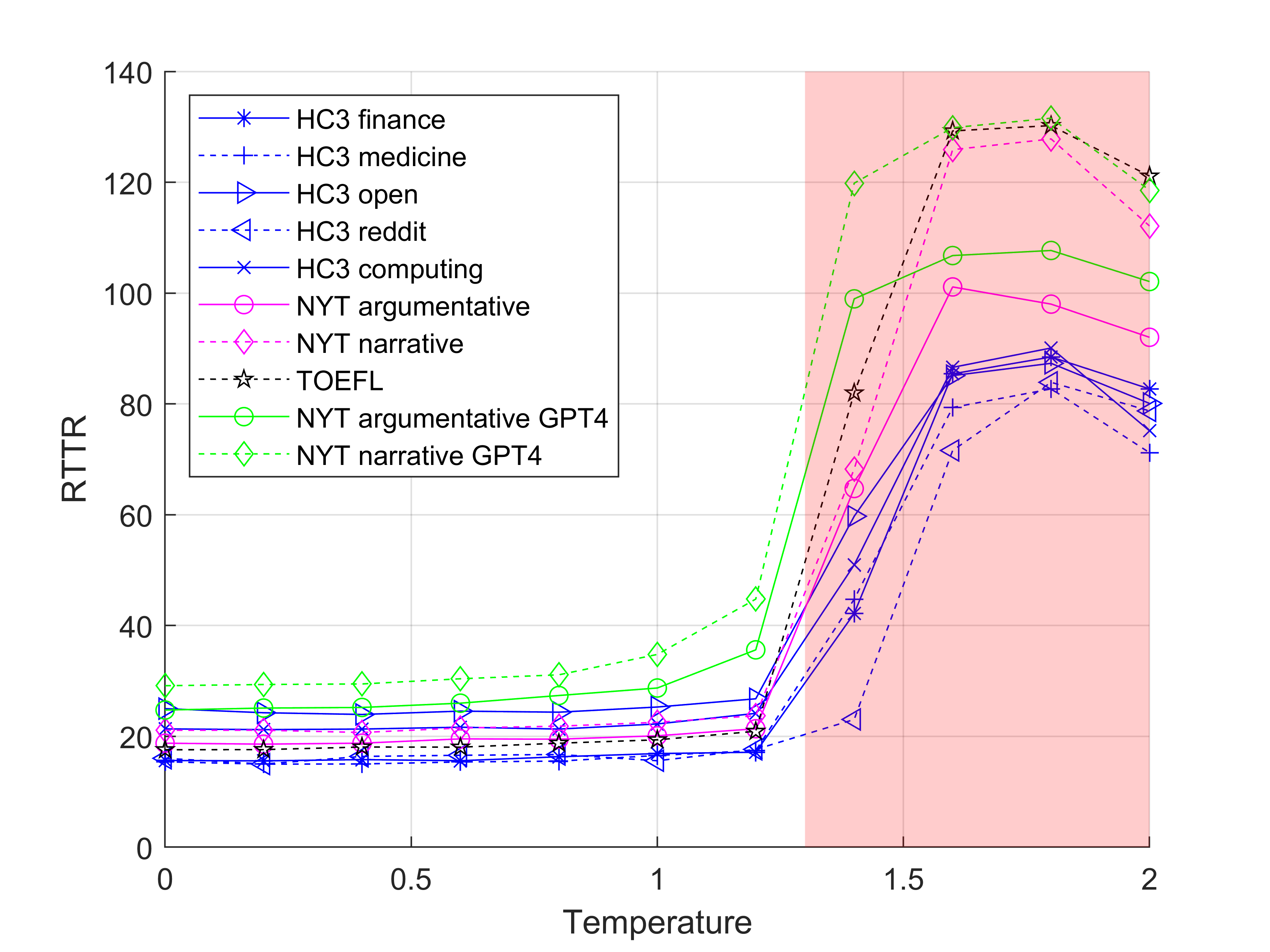}
  \caption{Lexical \textcolor{black}{diversity} global metrics (Maas, RTTR) versus Temperature, range values that produce incorrect results are marked in red.}
  \label{fig:T2}
\end{figure}

\begin{figure}[h]
  \centering
  \includegraphics[scale=0.45]{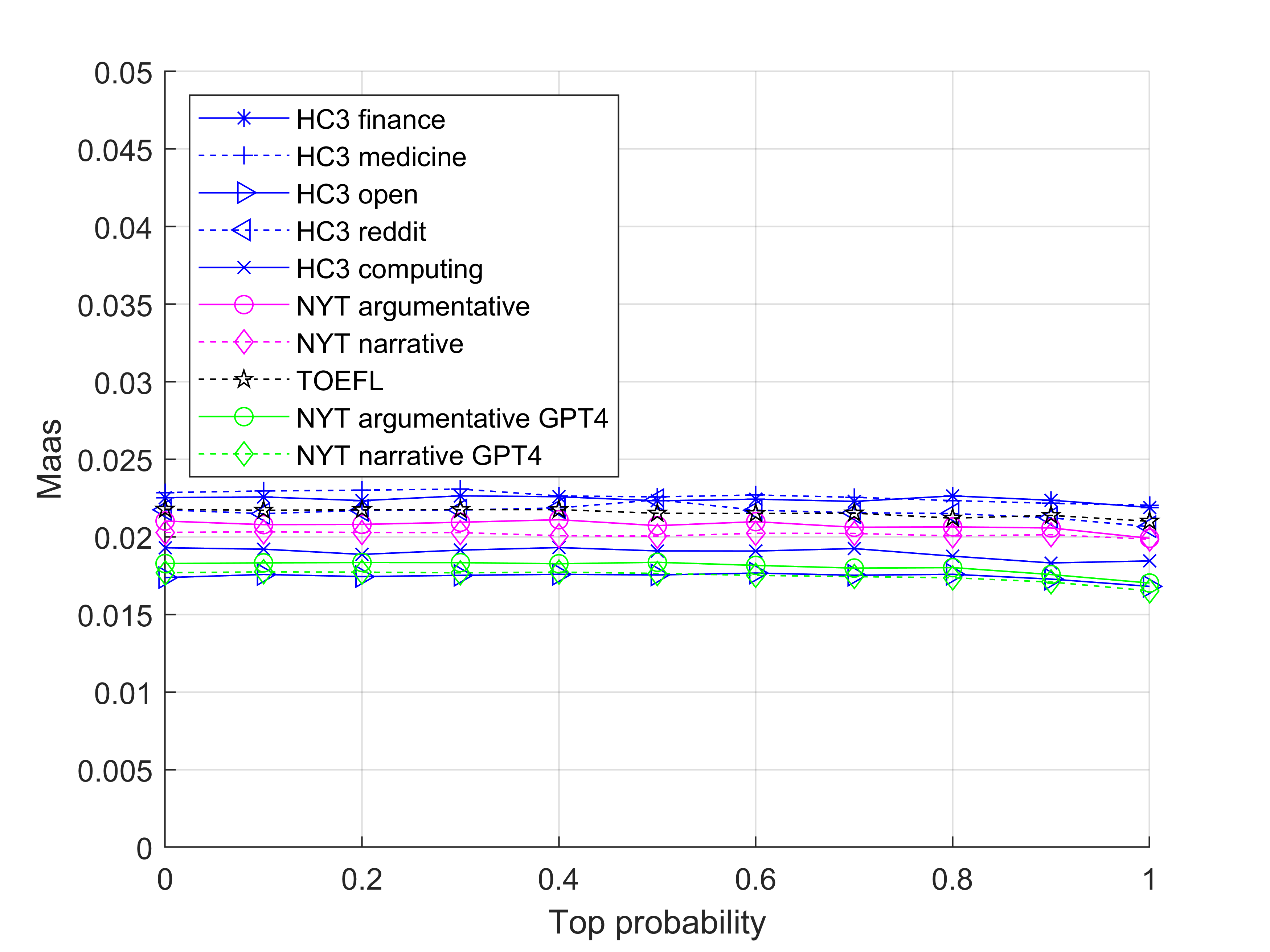}
  \includegraphics[scale=0.45]{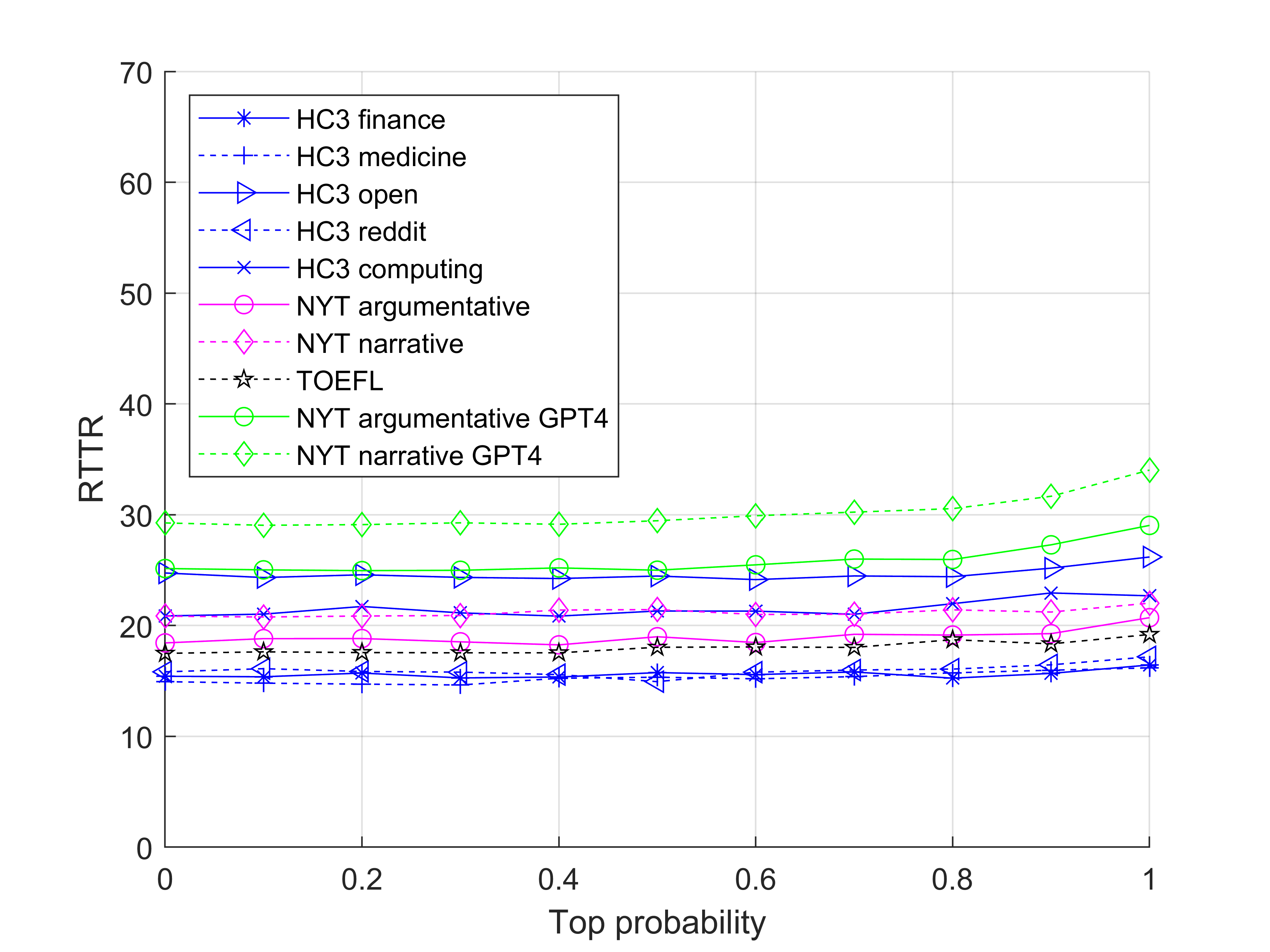}
  \caption{Lexical \textcolor{black}{diversity} global metrics (Maas, RTTR) versus Top Probability.}
  \label{fig:TP2}
\end{figure}

\begin{figure}[h]
  \centering
  \includegraphics[scale=0.45]{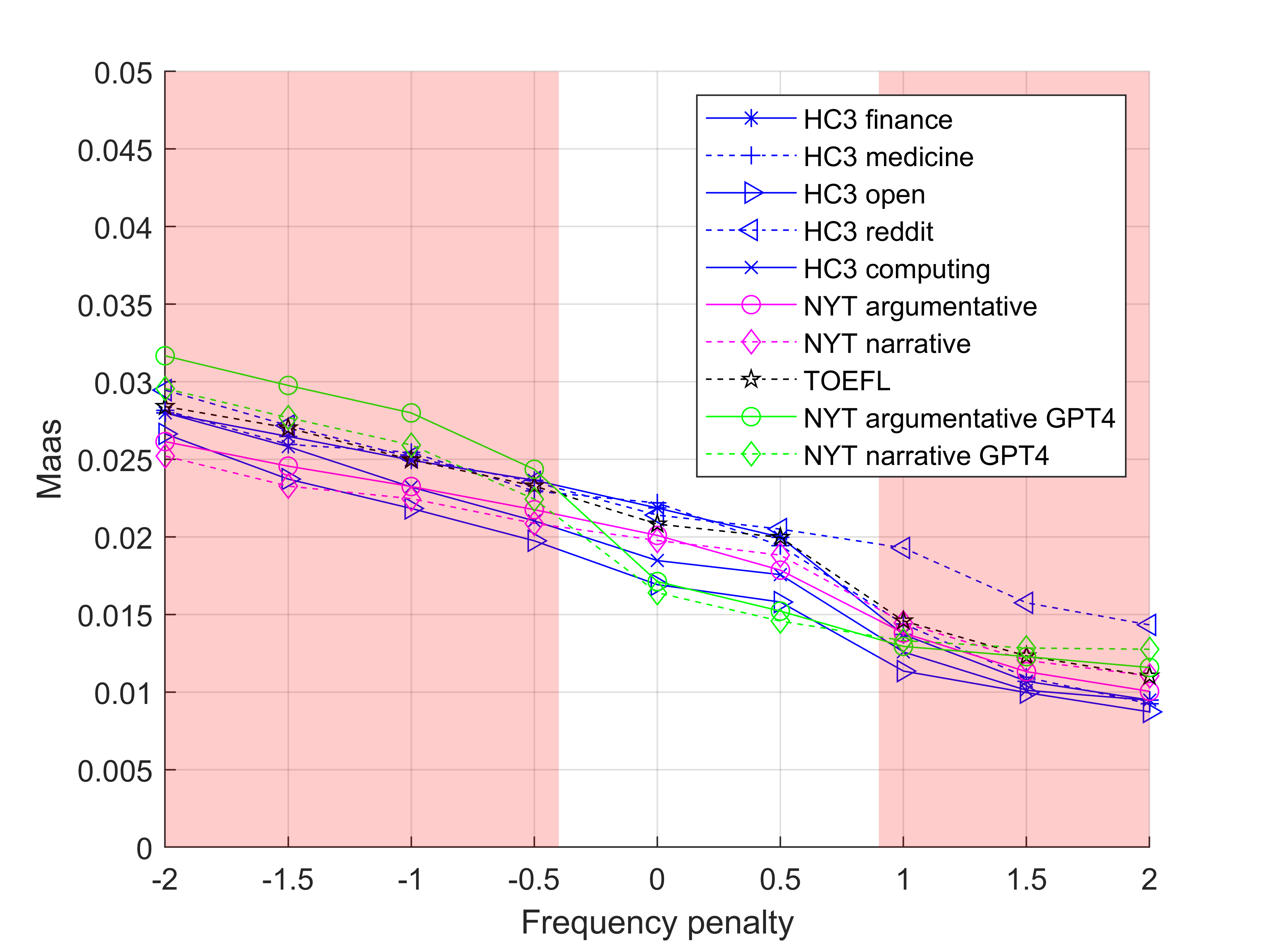}
  \includegraphics[scale=0.45]{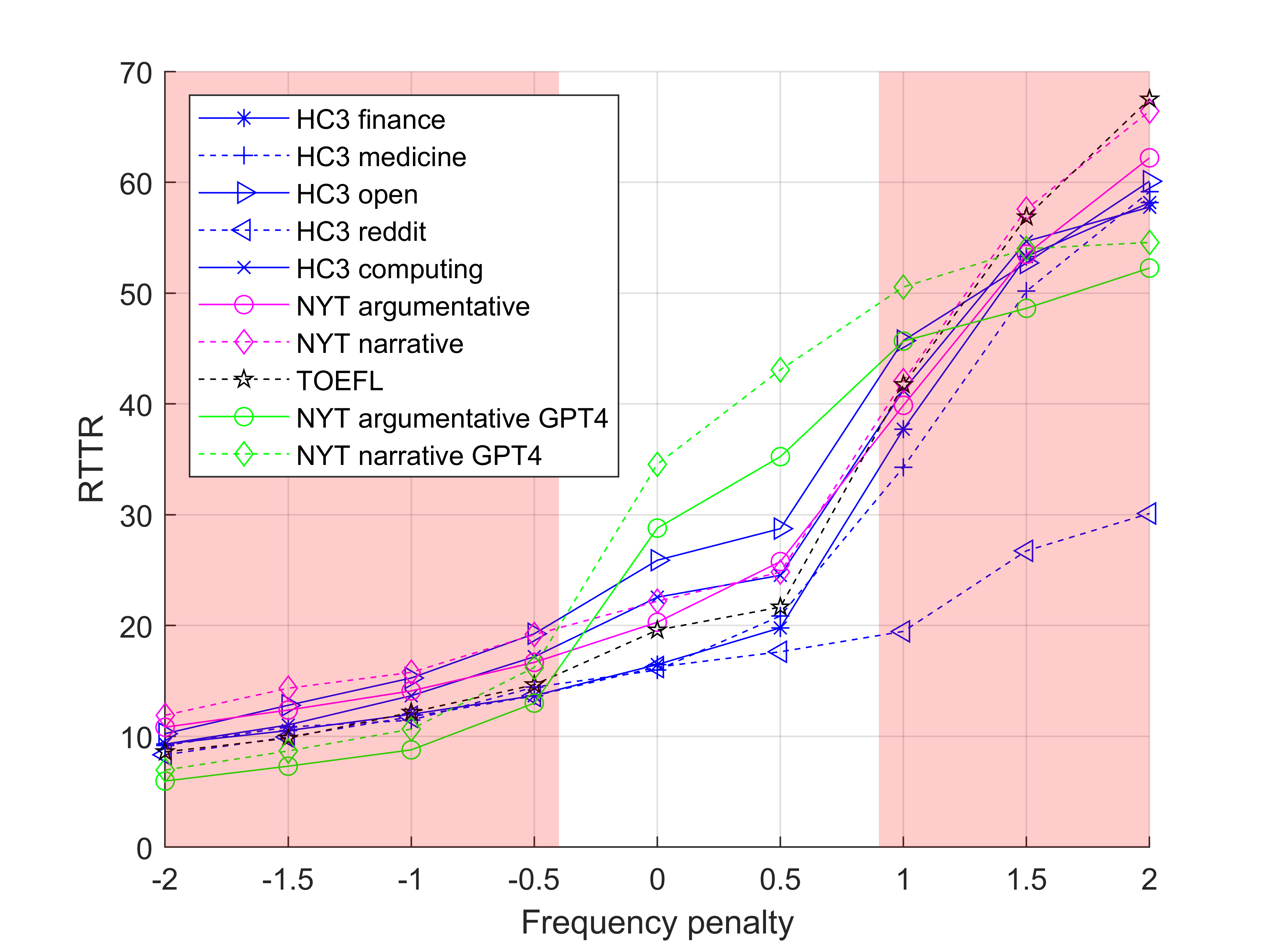}
  \caption{Lexical \textcolor{black}{diversity} global metrics (Maas, RTTR) versus Frequency Penalty, range values that produce incorrect results are marked in red.}
  \label{fig:F2}
\end{figure}

\begin{figure}[h]
  \centering 
  \includegraphics[scale=0.45]{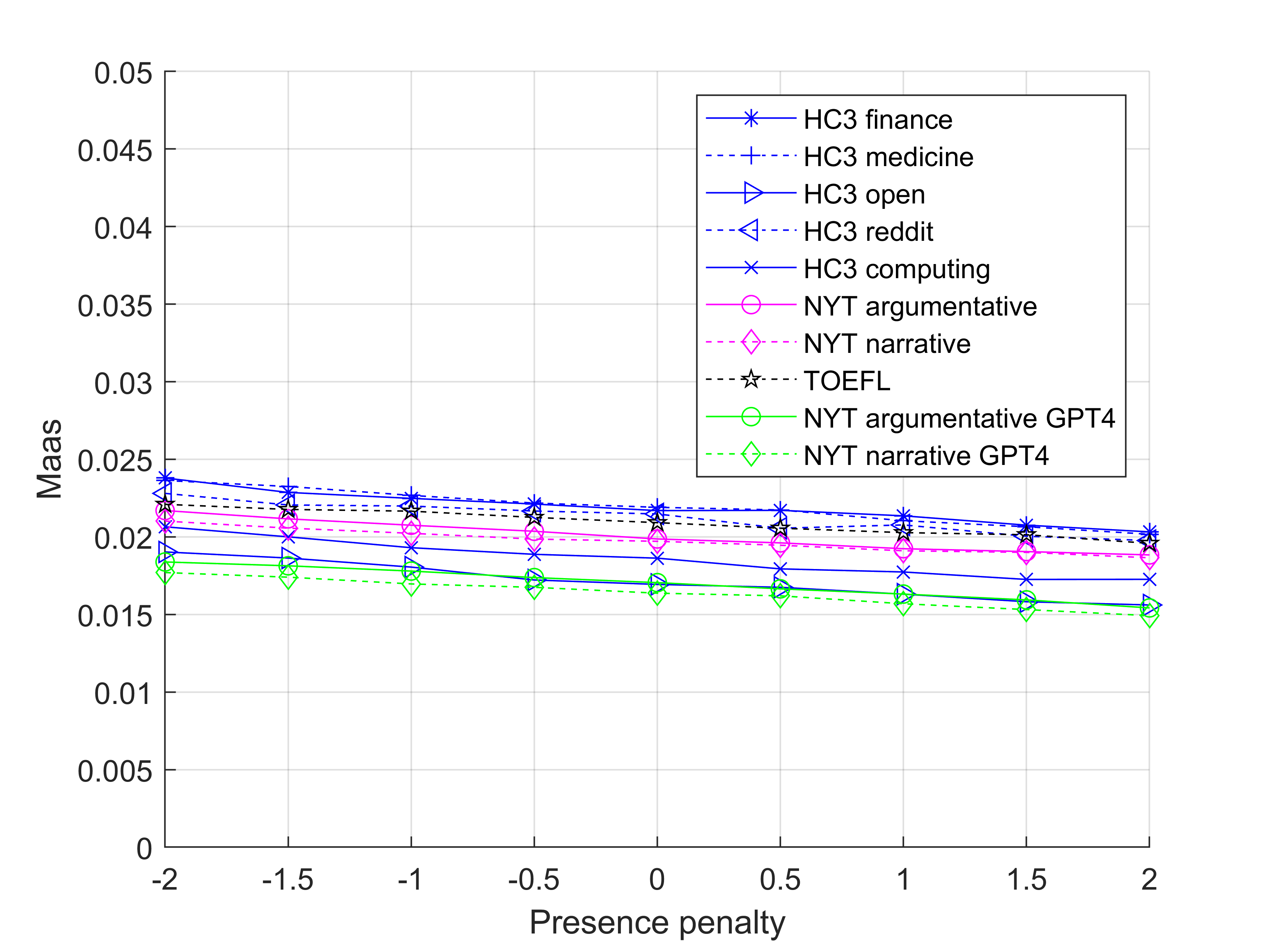}
  \includegraphics[scale=0.45]{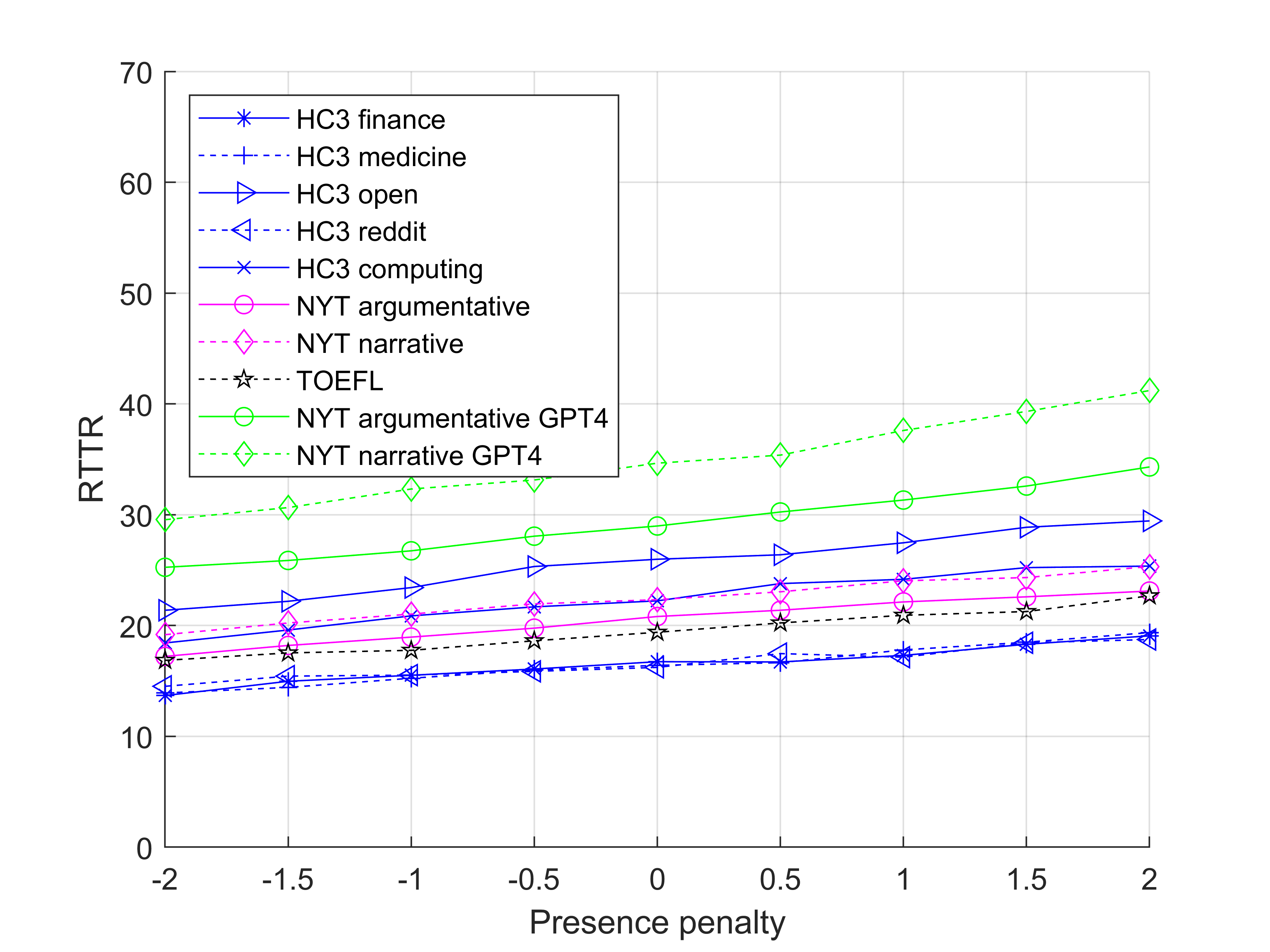}
  \caption{Lexical \textcolor{black}{diversity} global metrics (Maas, RTTR) versus Presence Penalty.}
  \label{fig:P2}
\end{figure}

The results for the second pair of metrics and the four parameters of the model are summarized in figures \ref{fig:T2},\ref{fig:TP2},\ref{fig:F2},\ref{fig:P2}. Note that in the case of the Maas metric, more lexical \textcolor{black}{diversity} corresponds to lower values. It can be seen that in general, they are consistent with the other two metrics. The results for temperature and frequency penalty show large differences in the regions where the output text is often invalid and a small increase with the parameter value in the rest of the range. The top probability does not have any significant impact on the metrics, only a small increase for values close to 1.0. Finally, the presence penalty increases the lexical \textcolor{black}{diversity} with larger (smaller) values of the RTTR (Maas) metrics. As for MATTR and MTLD, the trends are consistent across all the tasks and categories considered and for both ChatGPT3.5 and ChatGPT4. Comparing the lexical \textcolor{black}{diversity} of ChatGPT3.5 and ChatGPT4, for the same settings, ChatGPT4 has a larger richness in terms of diversity in most cases as for the other two metrics.

\begin{figure}[h]
  \centering 
  \includegraphics[scale=0.45]{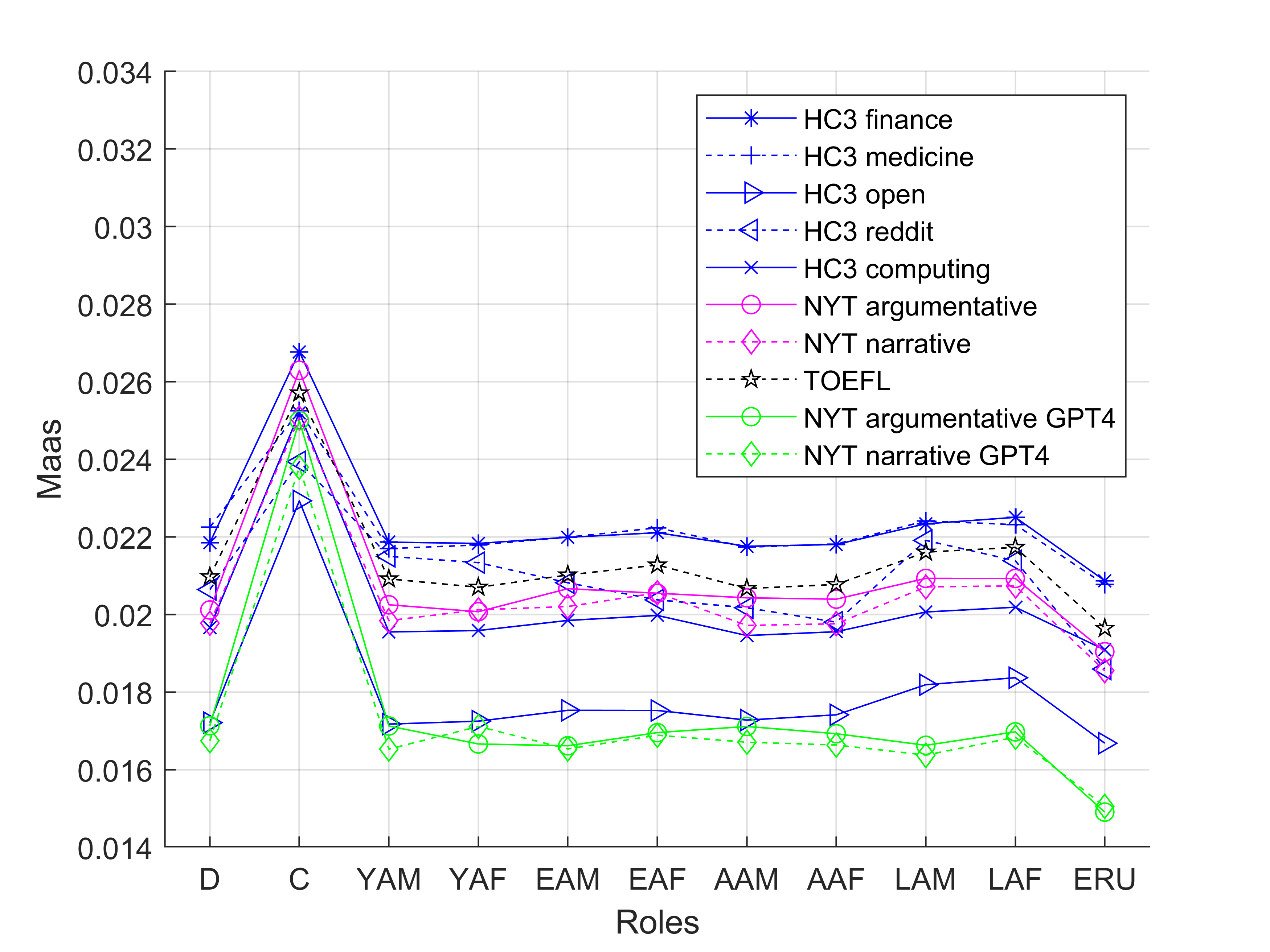}
  \includegraphics[scale=0.45]{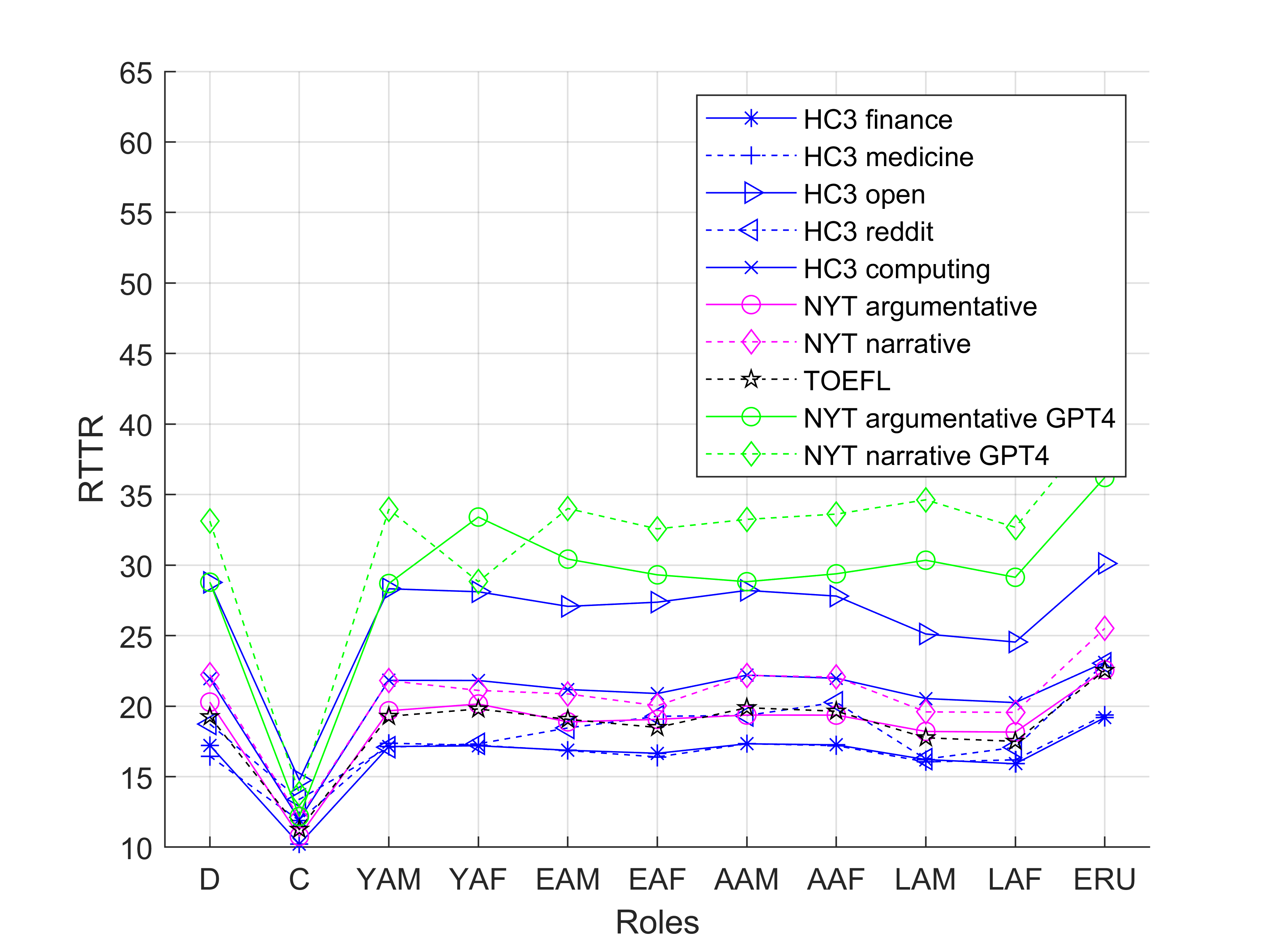}
  \caption{Lexical \textcolor{black}{diversity} global metrics (Maas, RTTR) for different roles.}
  \label{fig:R2}
\end{figure}

The results for the different roles on the second pair of metrics are shown in Figure \ref{fig:R2}. It can be observed that lexical \textcolor{black}{diversity} is lower for the ``child'' role as in the first two metrics. Additionally, it is also higher for ``erudite'' which was not the case in MATTR and MTLD. For the rest of the roles, differences are small and there is no clear trend. Therefore, age, sex, and social class do not seem to have a significant impact on lexical \textcolor{black}{diversity}. This means that in terms of lexical \textcolor{black}{diversity}, there seems to be no reason to be concerned about gender bias in ChatGPT \cite{ChatGPTBias}. The results for the child role may be due to the common use in the training of the models of datasets like ELI5 (Explain Like I am 5 years old) in which the questions ask for the answer to be written for a child \cite{fan2019eli5}.

\textcolor{black}{To understand if the results obtained for OpenAI models could be extrapolated to other LLMs, a last experiment was done. The tests were run for the different roles on two open LLMs, Mistral-7B \cite{Mistral} and LLaMa3-8B\footnote{\url{https://ai.meta.com/blog/meta-llama-3/}}. LLaMa3-8B also had lower lexical diversity for the child role with smaller variations for other roles, instead for Mistral, there was no significant variation even for the case of the child role. This suggests that further evaluation is needed to understand how lexical diversity varies in each LLM\footnote{The texts produced by Mistral-7B and LLaMa3-8B are also available in the public repository for further analysis.}. It would also be interesting to study the lexical features in languages other than English for which LLMs are known to perform worse.  }

\subsection{\textcolor{black}{Analysis and discussion}}

The analysis of the results can be summarized in the following insights:

\begin{enumerate}
    \item ChatGPT4 produces texts with larger lexical \textcolor{black}{diversity} than ChatGPT3.5. 
    \item The temperature and frequency penalty parameters produce invalid texts for some ranges of values that are excluded from the analysis.
    \item Temperature: lexical \textcolor{black}{diversity} increases slightly with temperature.
    \item Frequency penalty: lexical \textcolor{black}{diversity} increases slightly with frequency penalty.
    \item Presence penalty: lexical \textcolor{black}{diversity} increases with the presence penalty.
    \item Top probability: has no effect on lexical \textcolor{black}{diversity} unless for a small increase for values close to 1.0.
    \item Role: Age, sex, or social class seem to have little impact on lexical \textcolor{black}{diversity} except when the age is that of a child. 
    \item Role: asking the model explicitly to use a rich vocabulary (Erudite) increases the global lexical \textcolor{black}{diversity} metrics (RTTR and Mass) but not so much on the local ones (MATTR and MTLD). 
    \item Task: lexical \textcolor{black}{diversity} is generally larger for essay writing (TOEFL, NYT) than for question answering (HC3),
\end{enumerate}

These insights provide users with criteria to select the values of the model parameters to control the lexical \textcolor{black}{diversity} of the generated text and also to understand how parameter values selected for other reasons will impact lexical \textcolor{black}{diversity}. This illustrates the usefulness of carrying out the lexical \textcolor{black}{diversity} evaluation of LLMs. It is also important to note that our evaluation has focused on ChatGPT when used to answer questions and write essays on different topics. In other applications and uses, lexical \textcolor{black}{diversity} may show other dependencies with the model parameters. 

\textcolor{black}{Finally, it would be interesting to link these findings to the data used to train the LLMs. For example, understanding if the presence of different linguistic variations on the training set influences the results. Unfortunately, that is not possible as the training datasets are not publicly available. The same reasoning applies if we try to provide insights into how to improve the results. Intuitively a well-balanced dataset with high lexical diversity would improve the results, but no evidence can be provided to support that intuition. }


\section{Conclusion}

This work has evaluated the impact on the lexical \textcolor{black}{diversity} of the text generated by ChatGPT of the main model parameters such as temperature and top probability and also of assigning different roles to the model. To that end, a dataset has been created that exercises the models on different tasks and topics. This dataset has then been used to evaluate ChatGPT3.5 and ChatGPT4. The results show that lexical \textcolor{black}{diversity} increases with the presence penalty. This is reasonable as the presence penalty is a parameter that prevents the model from using words that have already been used. The results for other parameters that could be expected to also impact lexical \textcolor{black}{diversity} such as the frequency penalty or the temperature are limited as invalid texts are generated over wide ranges of the possible values. The top probability that restricts the model choices to the token with the highest scores does not seem to have a significant impact on the lexical \textcolor{black}{diversity} except for values close to one. Considering the roles, assigning a role to the model does not seem to affect its lexical \textcolor{black}{diversity} in terms of sex, age, and social class except when the model is asked to behave as a child. The results when the model is asked to behave as an erudite are mixed, it increases the lexical \textcolor{black}{diversity} according to two of the metrics and it does not for the other two metrics. Finally, lexical \textcolor{black}{diversity} tends to be larger for essay writing than for question answering.

The dataset and evaluation methodology presented in this paper is intended to facilitate further research on the lexical \textcolor{black}{diversity} of LLMs. The prompts used as well as the texts generated are available in a public repository so that they can be used to analyze other linguistic features. Future work can explore the lexical \textcolor{black}{diversity} of other LLMs. This would enable a better understanding of the impact of the model parameters on lexical \textcolor{black}{diversity} and also a comparison of different LLMs in terms of their use of the lexicon.

\begin{acks}

This work was supported by the FUN4DATE (PID2022-136684OB-C21/C22) project funded by the Spanish Agencia Estatal de Investigacion (AEI) 10.13039/501100011033, the Spanish project ENTRUDIT (grant no. TED2021-130118B-I00) and by the OpenAI research access program.

\end{acks}


\bibliographystyle{ACM-Reference-Format}

\bibliography{verbi}

\end{document}